\documentclass[10pt,journal,compsoc]{IEEEtran}

\ifCLASSOPTIONcompsoc
  \usepackage[nocompress]{cite}
\else
  \usepackage{cite}
\fi

\usepackage{graphicx}
\usepackage{subfigure}
\usepackage[numbers,square]{natbib}

\usepackage{multirow}
\usepackage{rotating}
\usepackage{paralist}
\usepackage{hhline}
\usepackage{amsmath}
\usepackage{amssymb}
\usepackage{amsfonts}
\usepackage{color}
\usepackage{wrapfig}
\usepackage{float}
\usepackage{gensymb}
\usepackage{array}
\usepackage{hyperref}

\usepackage{color}
\definecolor{dgreen}{rgb}{0,.7,0}
\definecolor{dyellow}{rgb}{.7,.7,0}
\definecolor{dred}{rgb}{.7,0,0}
\definecolor{dblue}{rgb}{0,0,0.7}
\definecolor{dmag}{rgb}{.6,0,0.6}
\definecolor{alexey}{rgb}{0.7,0,1}

\newcommand{\RGB}{\mathbf{x}}
\newcommand{\segm}{\mathbf{s}}
\newcommand{\class}{\mathbf{c}}
\newcommand{\view}{\mathbf{v}}

\newcommand{\augparams}{{\boldsymbol{\theta}}}
\newcommand{\vaeparams}{{\boldsymbol{\phi}}}
\newcommand{\weights}{\mathbf{W}}
\newcommand{\net}{g}
\newcommand{\fc}{h}
\newcommand{\deconv}{u}
\newcommand{\etal}{et al. }
\newcommand{\eg}{e.g. }
\newcommand{\loss}{L}
\newcommand{\bz}{\mathbf{z}}
\newcommand{\bSigma}{\mathbf{\Sigma}}
\newcommand{\bI}{\mathbf{I}}

\newcolumntype{C}{ >{\centering\arraybackslash} m{0cm} }
\newcolumntype{D}{ >{\centering\arraybackslash} m{9cm} }

\graphicspath{{./images/}}

\begin{document}
%
\title{Learning to Generate Chairs, Tables and Cars with Convolutional Networks }
%
%

\author{Alexey~Dosovitskiy,
        Jost~Tobias~Springenberg,
        Maxim~Tatarchenko,
        Thomas~Brox
\IEEEcompsocitemizethanks{\IEEEcompsocthanksitem All authors are with the Computer Science Department\protect\\ at the University of Freiburg\protect\\
E-mail: \{dosovits, springj, tatarchm, brox\}@cs.uni-freiburg.de}
\thanks{}}

%
%

\markboth{IEEE Transactions on Pattern Analysis and Machine Intelligence}%
{Dosovitskiy et al.: Learning to Generate Chairs, Tables and Cars with Convolutional Networks}
%

\IEEEtitleabstractindextext{%
\begin{abstract}
We train generative 'up-convolutional' neural networks which are able to generate images of objects given object style, viewpoint, and color.
We train the networks on rendered 3D models of chairs, tables, and cars.
Our experiments show that the networks do not merely learn all images by heart, but rather find a meaningful representation of 3D models allowing them to assess the similarity of different models, interpolate between given views to generate the missing ones, extrapolate views, and invent new objects not present in the training set by recombining training instances, or even two different object classes.
Moreover, we show that such generative networks can be used to find correspondences between different objects from the dataset, outperforming existing approaches on this task.
\end{abstract}

}

\maketitle

\IEEEdisplaynontitleabstractindextext

%
\IEEEpeerreviewmaketitle

\IEEEraisesectionheading{\section{Introduction}}

Generative modeling of natural images is a long standing and difficult task.
The problem naturally falls into two components: learning the distribution from which the images are to be generated, and learning the generator which produces an image conditioned on a vector from this distribution.
In this paper we approach the second sub-problem.
We suppose we are given high-level descriptions of a set of images, and only train the generator.
We propose to use an 'up-convolutional' generative network for this task and show that it is capable of generating realistic images.

In recent years, convolutional neural networks (CNNs, ConvNets) have become a method of choice in many areas of computer vision, especially on recognition~\cite{Krizhevsky_NIPS2012}.
Recognition can be posed as a supervised learning problem and ConvNets are known to perform well given a large enough labeled dataset.
In this work, we stick with supervised training, but we turn the standard classification CNN upside down and use it to generate images given high-level information.
This way, instead of learning a mapping from raw sensor inputs to a condensed, abstract representation, such as object identity or position, we generate images from their high-level descriptions.

Given a set of 3D models (of chairs, tables, or cars), we train a neural network capable of generating 2D projections of the models given the model number (defining the style), viewpoint, and, optionally, additional transformation parameters such as color, brightness, saturation, zoom, etc.
Our generative networks accept as input these high-level values and produce RGB images.
We train them with standard backpropagation to minimize the Euclidean reconstruction error of the generated image.

A large enough neural network can learn to perform perfectly on the training set.
That is, a network potentially could just learn by heart all examples and generate these perfectly, but would fail to produce reasonable results when confronted with inputs it has not seen during training.
We show that the networks we train do generalize to previously unseen data in various ways. Namely, we show that these networks are capable of: 1) knowledge transfer within an object class: given limited number of views of a chair, the network can use the knowledge learned from other chairs to infer the remaining viewpoints; 2) knowledge transfer between classes (chairs and tables): the network can transfer the knowledge about views from tables to chairs; 3) feature arithmetics: addition and subtraction of feature vectors leads to interpretable results in the image space; 4) interpolation between different objects within a class and between classes; 5) randomly generating new object styles.


After a review of related work in Section~\ref{sec:related_work}, we describe the network architecture and training process in Section~\ref{sec:model}. In Section~\ref{sec:train_params} we compare different network architectures and dataset sizes, then in Section~\ref{sec:experiments} we test the generalization abilities of the networks and apply them to a practical task of finding correspondences between different objects. Finally, in Section~\ref{sec:analysis} we analyze the internal representation of the networks. 

\section{Related work} \label{sec:related_work}
Work on generative models of images typically addresses the problem of unsupervised
learning of a data model which can generate samples
from a latent representation.
Prominent examples from this line of work are restricted Boltzmann
machines (RBMs) \cite{HintonSalakhutdinov2006} and Deep Boltzmann Machines
(DBMs) \cite{Salakhutdinov_2009}, as well as the plethora of models derived from
them~\cite{Hinton_NC2006, Memisevic07, Lee_ICML2009, Tang_AISTATS2012, Ranzato_CVPR2011, WuCVPR15}. RBMs and DBMs are undirected graphical models which aim to build
a probabilistic model of the data and treat encoding and generation as
an (intractable) joint inference problem.
Most related to our approach are Convolutional Deep Belief Networks (CDBNs) of Lee \etal~\cite{Lee_ICML2009} making use of ``unpooling'' and ShapeNets of Wu \etal~\cite{WuCVPR15} training a 3D variant of CDBN to generate 3D models of furniture.


A different approach is to train directed graphical models of the data
distribution. This includes a wide variety of methods ranging
from Gaussian mixture models~\cite{Francos_GMM03,Theis_2012} to autoregressive models \cite{Larochelle_2011}
and stochastic variations of neural networks
\cite{Bengio_ICML2014,Goodfellow_NIPS2014,Rezende_ICML2014,Kingma_ICLR2014,Tang_NIPS2013}.
Among them Rezende \etal~\cite{Rezende_ICML2014}
developed an approach for training a generative model with variational inference by performing
(stochastic) backpropagation through a latent Gaussian representation.
The generative adversarial networks approach presented in Goodfellow \etal \cite{Goodfellow_NIPS2014} models natural images using a ''deconvolutional''
generative network that is similar to our architecture.

Most unsupervised generative models can be extended to incorporate
label information, forming semi-supervised and conditional generative
models which lie between fully unsupervised approaches and our work.
Examples
include: gated conditional RBMs
\cite{Memisevic07} for modeling image transformations, training RBMs
to disentangle face identity and pose information using conditional
RBMs \cite{Reed_icml2014}, and learning a generative model of digits conditioned
on digit class using variational
autoencoders~\cite{Kingma_NIPS2014}. In contrast to our work, these
approaches are typically restricted to small models and images,
and they often require an expensive inference procedure -- both
during training and for generating images.

Since the publication of the conference paper this work
  is based on \cite{DosovitskiyCVPR15}, there has been a resurgence in
  research on using neural networks for generating images. This line
  of work includes an increasing amount of papers using large
  ``up-convolutional'' neural networks for modelling realistic images
  using architectures that are similar to -- or are derived from --
  the ones presented in this paper
  \cite{KulkarniNIPS15,ReedNIPS15,DentonNIPS15,Radford2015}. Denton \etal~\cite{DentonNIPS15} and Radford
  \etal \cite{Radford2015} train conditional convolutional generative models via the adversarial networks approach.
  These networks are capable of generating high-fidelity natural images.
  Reed \etal~\cite{ReedNIPS15} learn to generate animations of computer game characters based on a learned code that represents image transformations.

The general difference of our approach to prior work on learning
generative models is that we
assume a high-level latent representation of the images is
given and use supervised training.
This allows us 1) to generate relatively large high-quality images of $128 \times 128$ pixels (as compared to
maximum of $48 \times 48$ pixels in most previous works)
and 2) to completely control which images to generate rather
than relying on random sampling.
The downside is, of course, the need for a label that fully describes the appearance of each
image. 

This requirement is slightly relaxed by Kulkarni \etal~\cite{KulkarniNIPS15} who learn to generate images of objects, assuming only partial knowledge about the parameters of the scene (namely, which parameters are changing and which are fixed). 



Modeling of viewpoint variation is often considered in the context of pose-invariant face recognition~\cite{Blanz_PAMI2003}.
In a recent work Zhu \etal~\cite{Zhu_NIPS2014} approached this task with a neural network: their network takes a face image
as input and generates a random view of this
face together with the corresponding viewpoint.
The network is fully connected and hence restricted to small
 images and, similarly to generative models, requires random sampling to generate a desired view. This makes it inapplicable to modeling large and diverse images, such as the chair images we model.


Our work is also loosely related to applications of CNNs to non-discriminative tasks, such as
super-resolution~\cite{Dong_ECCV2014} or inferring depth from a single image~\cite{Eigen_NIPS2014}.

\begin{figure*}
\begin{center}
   \includegraphics[width=0.95\linewidth]{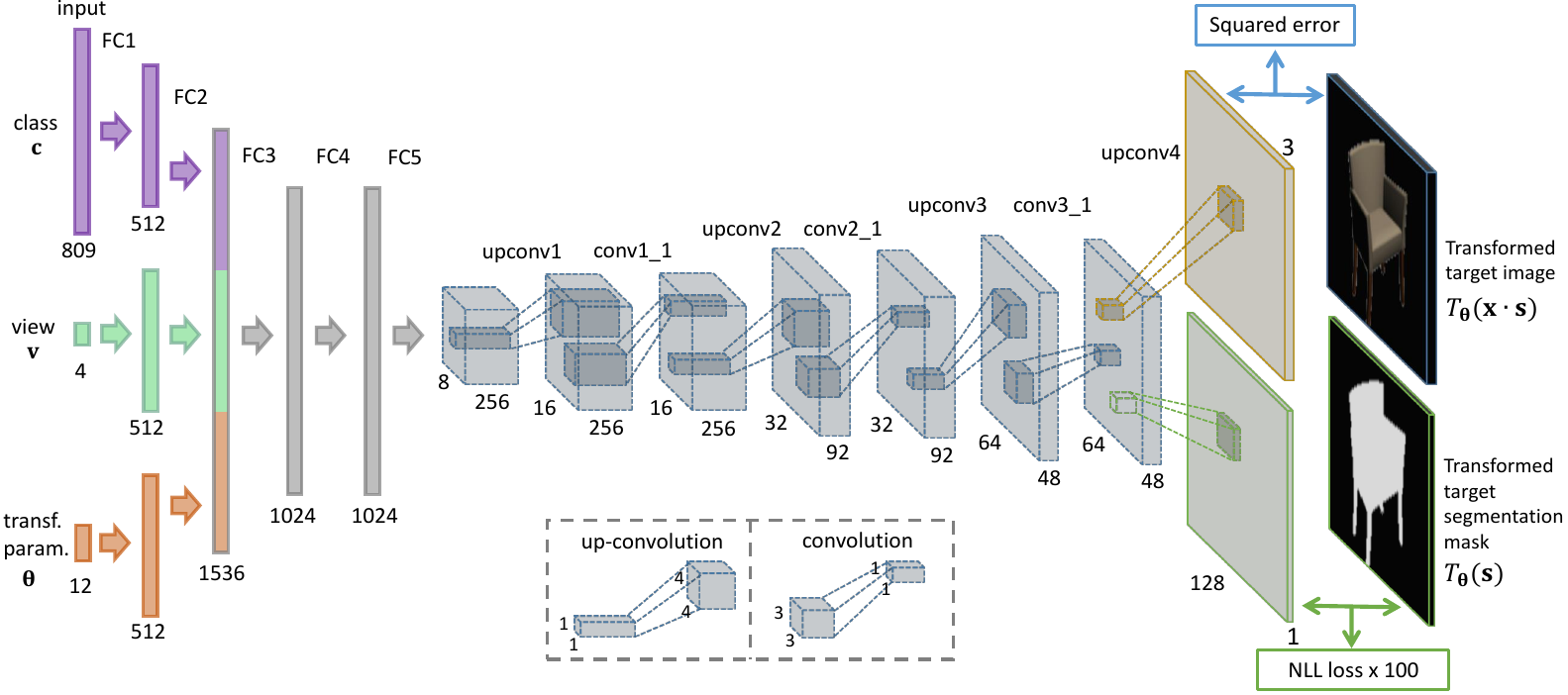}
\end{center}
\vspace*{-0.2cm}
   \caption{Architecture of a 1-stream deep network (``1s-S-deep``) that generates $128 \times 128$ pixel images. Layer names are shown above: FC - fully connected, upconv - upsampling+convolution, conv - convolution.}
   \vspace{-0.3cm}
\label{fig:network_architecture}
\end{figure*}

\section{Model description} \label{sec:model}


Our goal is to train a neural network to generate accurate images of objects from a high-level description: style, orientation with respect to the camera, and additional parameters such as color, brightness, etc.
The task at hand hence is the inverse of a typical recognition task: rather than converting an image to a compressed high-level representation, we need to generate an image given the high-level parameters.

Formally, we assume that we are given a dataset of examples $D = \lbrace (\class^1, \view^1, \augparams^1), \dots ,(\class^N, \view^N, \augparams^N) \rbrace$ with targets $O = \lbrace (\RGB^1, \segm^1), \dots, (\RGB^N, \segm^N) \rbrace$. The input tuples consist of three vectors: $\class$ is the one-hot encoding of the model identity (defining the style), $\view$~-- azimuth and elevation of the camera position (represented by their sine and cosine~\footnote{We do this to deal with periodicity of the angle. If we simply used the number of degrees, the network would have no way to understand that 0 and 359 degrees are in fact very close.}) and $\augparams$~-- the parameters of additional artificial transformations applied to the images. The targets are the RGB output image $\RGB$ and the segmentation mask $\segm$. We note that predicting segmentation masks is not a strict requirement for good generative capabilities but it allows us to easily separate the generated images from the background -- and potentially replace it. In most figures we use these predicted segmentation masks to replace black background with white.
 
We include artificial transformations $T_\augparams$ described by the randomly generated parameter vector $\augparams$ to increase the amount of variation in the training data and reduce overfitting, analogous to data augmentation in discriminative CNN training~\cite{Krizhevsky_NIPS2012}.
Each $T_\augparams$ is a combination of the following transformations: in-plane rotation (up to $\pm 12\degree$), translation (up to $\pm10 \%$ of image size), zoom-in ($100\%$ to $135\%$), stretching horizontally or vertically (up to $10\%$), changing hue (arbitrary random additive factor), changing saturation ($25\%$ to $400\%$), changing brightness ($35\%$ to $300\%$).

\subsection{Network architectures}
Conceptually the generative network, which we formally refer to as
$\net(\class, \view, \augparams)$, looks like a usual CNN turned upside down.
It can be thought of as the composition of two processing steps $\net = \deconv \circ \fc$.
We experimented with several architectures, one of them is shown in Figure~\ref{fig:network_architecture}.

Layers FC1 to FC4 first build a shared, high dimensional hidden representation
$\fc(\class, \view, \augparams)$ from the input parameters. Within
these layers the three input vectors are first independently fed through two fully connected layers with $512$ neurons each, and then the outputs of these three streams are concatenated. This independent processing is followed by two fully connected layers with $1024$ neurons each, yielding the response of the fourth fully connected layer (FC4).

The expanding part $\deconv$ of the network consists of layers FC5 and upconv1 to upconv4.
It generates an image $\deconv_{RGB}(h)$ and a segmentation mask $\deconv_{segm}(h)$ from the hidden representation $\fc$.
We experimented with two architectures for $\deconv$.
In one of them, as depicted in Figure~\ref{fig:network_architecture}, the image and the segmentation mask are generated based on a shared feature representation. 
A fully connected layer FC5 outputs a $16384$-dimensional vector which is reshaped to a $8 \times 8$ multichannel image and fed through $3$ ''upsampling+convolution`` layers with $4\times 4$ filters and $2 \times 2$ upsampling, each followed by a convolutional layer with $3 \times 3$ filters.
We found that adding a convolutional layer after each up-convolution significantly improves the quality of the generated images.
The last upconv4 layer predicts both the RGB image and the segmentation mask.
In an alternative 2-stream architecture, the network splits into two streams (predicting RGB image and the segmentation mask) right after the layer FC4.
We describe and compare different architectures in section~\ref{sec:arch_comparison}.

\begin{figure}[t]
\begin{center}
   \includegraphics[width=0.9\linewidth]{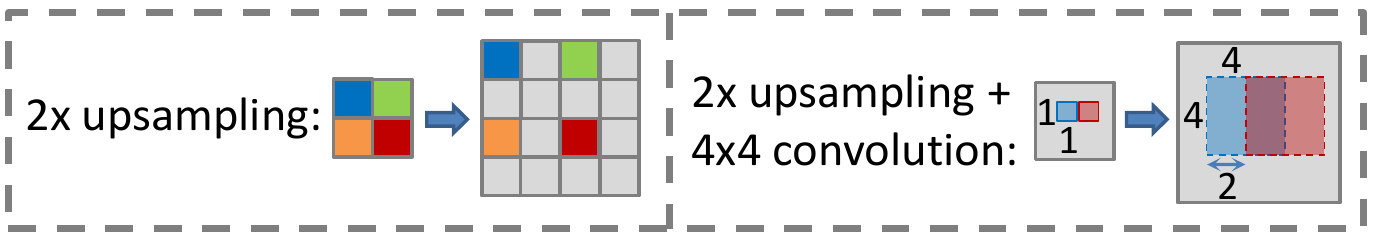}
\end{center}
   \caption{Illustration of upsampling (left) and upsampling+convolution (right) as used in the generative network.}
   \vspace{-0.2cm}
\label{fig:unpooling}
\end{figure}

In order to map the dense $8 \times 8$ representation to a high dimensional image, we need to unpool the feature maps (i.e. increase their spatial span) as opposed to the pooling (shrinking the feature maps) implemented by usual CNNs.
This is similar to the ``deconvolutional'' layers used in previous work \cite{Zeiler_ECCV2014,Goodfellow_NIPS2014,Zeiler_ICCV2011}.
As illustrated in Figure~\ref{fig:unpooling} (left), we perform unpooling by simple ``bed of nails'' upsampling, that is, replacing each entry of a feature map by an $s \times s$ block with the entry value in the top left corner and zeros elsewhere.
This increases the width and the height of the feature map $s$ times.
We used $s=2$ in our networks.
When a convolutional layer is preceded by such an upsampling operation we can think of upsampling+convolution (``up-convolution'') as the opposite of the convolution+pooling steps performed in a standard CNN, see Figure~\ref{fig:unpooling} right.
This figure also illustrates how in practice the upsampling and convolution steps do not have to be performed sequentially, but can be combined in a single operation.
Implementation-wise this operation is equivalent to a backward pass through a usual convolutional layer with stride $s$.

In all our networks each layer, except the output, is followed by a rectified linear (ReLU) nonlinearity.
In most experiments we generated $128 \times 128$ pixel images, but we also experimented with $64 \times 64$ and $256 \times 256$ pixel images.
The only difference in architecture in these cases is one less or one more up-convolution, respectively.

\subsection{Network training}
The network parameters $\weights$, consisting of all layer weights
and biases, are trained by minimizing the error of reconstructing the segmented-out chair image and the segmentation mask (the weights $\weights$ are omitted from the arguments of $\fc$ and $\deconv$ for brevity of notation):
\begin{equation}
  \begin{aligned}
        \min_{\weights}\;\; \sum_{i=1}^N  \qquad &\loss_{RGB} \left(T_{\augparams^i}(\RGB^i \cdot \segm^i),\, \deconv_{RGB}(\fc(\class^i, \view^i, \augparams^i))\right) \\
        + \lambda \cdot & \loss_{segm} \left(T_{\augparams^i}(\segm^i),\, \deconv_{segm}(\fc(\class^i, \view^i, \augparams^i))\right),
  \end{aligned}
\end{equation}
where $\loss_{RGB}$ and $\loss_{segm}$ are loss functions for the RGB image and for the segmentation mask respectively, and $\lambda$ is a weighting term, trading off between these two.
In our experiments $\loss_{RGB}$ was always squared Euclidean distance, while for $\loss_{segm}$ we tried two choices: squared Euclidean distance and negative log-likelihood loss preceded by a softmax layer.
We set $\lambda = 0.1$ in the first case and $\lambda = 100$ in the second case.


\begin{figure}
\begin{center}
   \includegraphics[width=1.\linewidth]{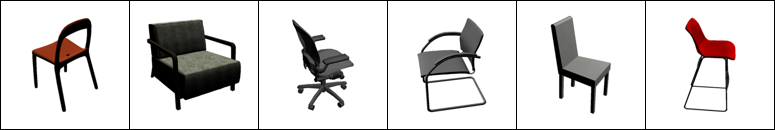}
   \includegraphics[width=1.\linewidth]{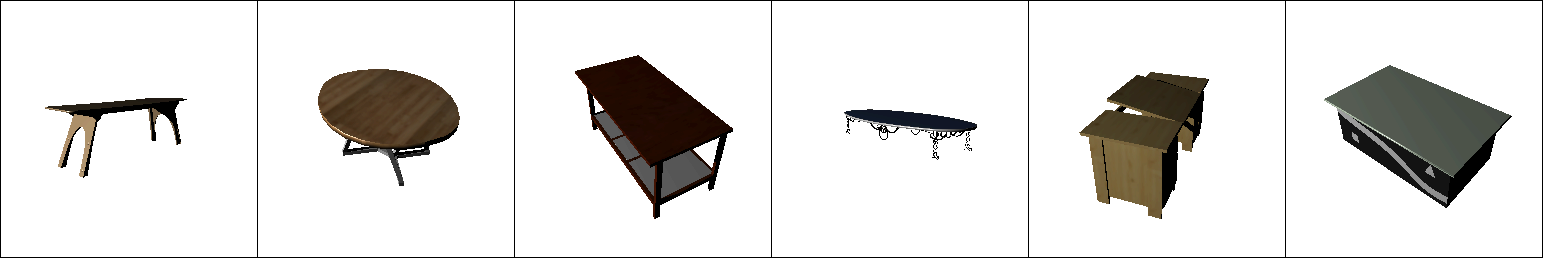}
   \includegraphics[width=1.\linewidth]{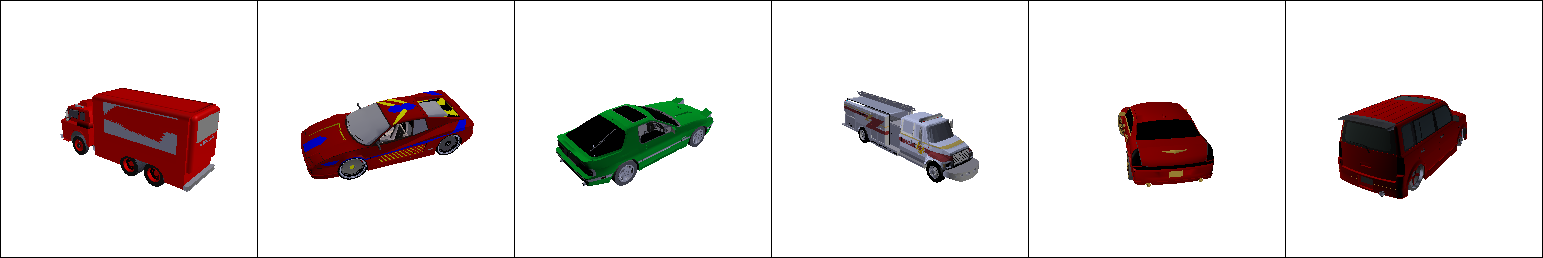}
\end{center}
   \caption{Representative images used for training the networks. \textbf{Top:} chairs, \textbf{middle:} tables, \textbf{bottom}: cars.}
\label{fig:example_chairs}
\end{figure}

\subsection{Probabilistic generative modeling}
\label{sec:prob_gen}
As we show below, the problem formulation described in the beginning of this section allows us to learn a generator network which can interpolate between different objects.
The learned network, however, implements a deterministic function from a high-level description (including the object identity as given by the one-hot encoding $\class$) to a generated image.
The shared structure between multiple objects is only implicitly learned by the network and not explicitly accessible.
It can be used to generate new ``random'' objects by blending between objects from the training set, but this heuristics comes with no guarantees regarding the quality of the generated images.
To get a more principled way of generating new objects we train a probabilistic generative model with an intermediate Gaussian representation.
We replace the independent processing stream for the class identity $\class$ in layer FC2 (the magenta part in Figure \ref{fig:network_architecture}) with random samples drawn from an inference network $q(\bz \mid \class) = \mathcal{N}(\mu_\bz, \bSigma_\bz)$ that learns to capture the underlying (latent) structure of different chair images.
This inference network can be learned alongside the generator described above by maximizing a variational bound on the sample log-likelihood.
A full description of the probabilistic generative model training is given in Appendix A.

\subsection{Datasets}
As training data for the generative networks we used renderings of 3D models of different objects: chairs, made public by Aubry \etal~\cite{Aubry_CVPR2014}, as well as car and table models from the ShapeNet~\cite{shapenet} dataset.

Aubry \etal provide renderings of 1393 aligned chair models, each rendered from 62 viewpoints: 31 azimuth angles (with a step of $11\degree$) and 2 elevation angles ($20\degree$ and $30\degree$), with a fixed distance to the chair. We found that the dataset includes many near-duplicate models, models differing only by color, or low-quality models. After removing these we ended up with a reduced dataset of 809 models, which we used in our experiments. We cropped the renders to have a small border around the chair and resized to a common size of $128 \times 128$ pixels, padding with white where necessary to keep the aspect ratio. Example images are shown in Figure~\ref{fig:example_chairs}. For training the network we also used segmentation masks of all training examples, which we produced by subtracting the monotonous white background.

We took models of cars and tables from ShapeNet, a dataset containing tens of thousands of consistently aligned 3D models of multiple classes.
We rendered a turntable of each model in Blender\footnote{\url{https://www.blender.org}} using 36 azimuth angles (from $0\degree$ to $350\degree$ with step of $10\degree$) and 5 elevation angles (from $0\degree$ to $40\degree$ with step of $10\degree$), which resulted in 180 images per model.
Positions of the camera and the light source were fixed during rendering.
For experiments in this paper we used renderings of 7124 car models and 1000 table models.
All renderings are $256 \times 256$ pixels, and we additionally directly rendered the corresponding segmentation masks.
Example renderings are shown in Figure~\ref{fig:example_chairs}.

\section{Training parameters} \label{sec:train_params}
In this section we describe the details of training, compare different network architectures and analyze the effect of dataset size and data augmentation on training.

\subsection{Training details}
For training the networks we built on top of the \emph{caffe} CNN implementation~\cite{caffe}. We used Adam~\cite{KingmaB14} with momentum parameters $\beta_1=0.9$ and $\beta_2=0.999$, and regularization parameter $\epsilon = 10^{-6}$.
Mini-batch size was $128$.
We started with a learning rate of $0.0005$ for $250,000$ mini-batch iterations, and then divided the learning rate by $2$ after every $100,000$ iterations, stopping after $500,000$ iterations.
We initialized the weights with Gaussian noise with variance computed based on the input dimensionality, as suggested by Susillo \cite{Sussillo14} and He \etal \cite{HeZR015}\, .

In most experiments we trained networks generating $128 \times 128$ pixel images.
In the viewpoint interpolation experiments in section~\ref{sec:angle_interpolation} we generated $64 \times 64$ pixel images to reduce the computation time, since multiple networks had to be trained for those experiments.
When working with cars, we tried generating $256 \times 256$ images to check if deeper networks capable of generating these larger images can be successfully trained.
We did not observe any complications during training of these deeper networks.

\subsection{Comparing network architectures}
\label{sec:arch_comparison}
As mentioned above, we experimented with different network architectures.
These include:
\begin{itemize}
\item[``1s-S-deep'']~-- a network shown in Figure~\ref{fig:network_architecture}, with one stream, negative log-likelihood (NLL) loss on the segmentation mask and convolutions between up-convolutions ;
\item[``1s-S'']~-- same as ``1s-S-deep'', but without convolutions between up-convolutions;
\item[``1s-S-wide'']~-- same as ``1s-S'', but with roughly $1.3$ times more channels in each up-convolutional layer;
\item[``2s-S'']~-- same as ``1s-S'', but with two separate streams for the RGB image and the segmentation mask;
\item[``2s-E'']~-- same as ``2s-S'', but with squared Euclidean loss on the segmentation mask.

\end{itemize}

\begin{figure}
\begin{center}
\small{
\quad\; GT \qquad\; 2s-E \qquad\; 2s-S \qquad\; 1s-S \quad\; 1s-S-wide \, 1s-S-deep }
\includegraphics[width=1\linewidth]{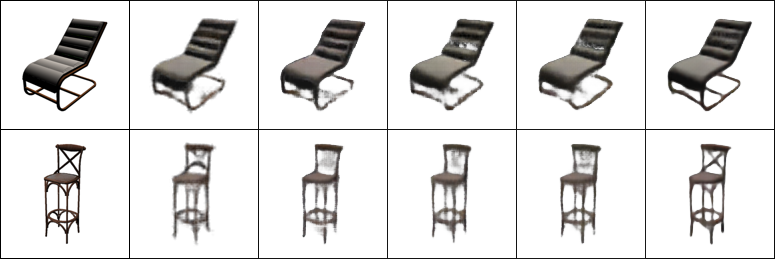}
\end{center}
   \caption{Qualitative results with different networks trained on chairs.
   See the description of architectures in section~\ref{sec:arch_comparison}.}
\label{fig:qual_arch_comparison}
\end{figure}

Images generated with these architectures, as well as the ground truth (GT),  are shown in Figure~\ref{fig:qual_arch_comparison}.
Reconstruction errors are shown in Table~\ref{tbl:quant_arch_comparison}.
Clearly, the deeper ``1s-S-deep'' network is significantly better than others both qualitatively and quantitatively.
For this reason we used this network in most experiments.

\subsection{Training set size and data augmentation}
\begin{figure}
\begin{center}
\quad GT \qquad\;\; 500 \qquad\;\; 1000 \qquad 3000 \qquad 7124 \quad 1000+aug
\includegraphics[width=1\linewidth]{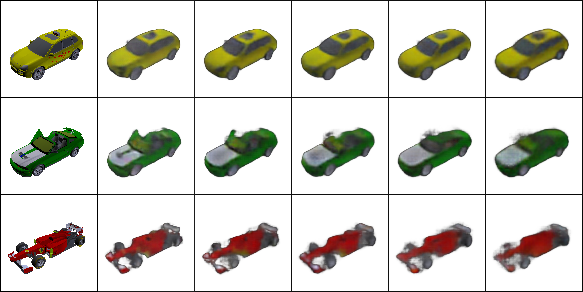}
\end{center}
   \caption{Qualitative results for different numbers of car models in the training set, without and with (rightmost column) data augmentation.}
\label{fig:qual_var_num_samples}
\end{figure}

We experimented with the training set size and analyzed what is the effect of data augmentation.
We used cars for these experiments, since we have more car models available.
While keeping the network architecture fixed, we varied the training set size.
Example generated images are shown in Figure~\ref{fig:qual_var_num_samples}.
Each column corresponds to a different number of models in the training set, and all networks except the one in the rightmost column were trained without data augmentation.
While for a standard car model (top row) there is not much difference, for difficult models (other rows) smaller training set leads to better reconstruction of fine details.
The effect of data augmentation is qualitatively very similar to increasing the training set size.
Reconstruction errors shown in Table~\ref{tbl:quant_var_num_samples} support these observations.

\begin{figure}
\begin{center}
\includegraphics[width=1\linewidth]{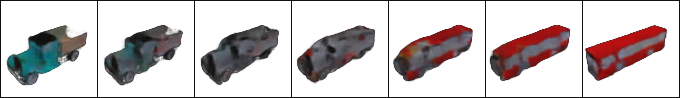}
\includegraphics[width=1\linewidth]{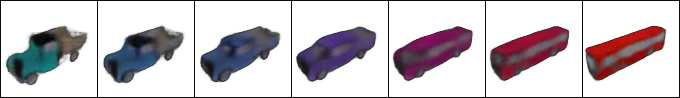}
\end{center}
   \caption{Interpolation between two car models.
   \textbf{Top:} without data augmentation, \textbf{bottom:} with data augmentation.}
\label{fig:aug_interpolation}
\end{figure}

Data augmentation leads to worse reconstruction of fine details, but it is expected to lead to better generalization.
To check this, we tried to morph one model into another by linearly interpolating between their one-hot input style vectors.
The result is shown in Figure~\ref{fig:aug_interpolation}.
Note how the network trained without augmentation (top row) better models the images from the training set, but fails to interpolate smoothly.

\section{Experiments} \label{sec:experiments}
We show how the networks successfully model the complex data and demonstrate their generalization abilities by generating images unseen during training: new viewpoints and object styles.
We also show an application of generative networks to finding correspondences between objects from the training set.

\begin{table}
\begin{center}
\begin{tabular}{l||c|c|c|c|c}
Net                   & 1s-S-deep & 1s-S   & 1s-S-wide & 2s-S     & 2s-E    \\
\hline
MSE ($\cdot 10^{-3}$) & $2.90$    & $3.51$ & $3.41$    & $3.44$   & $3.43 $  \\
\#param               & $19$M     & $18$M  & $23$M     & $27$M    & $27$M   
\end{tabular}
\end{center}
\caption{Per-pixel mean squared error of the generated chair images with different network architectures and the number of parameters in the expanding parts of these networks.
}
\label{tbl:quant_arch_comparison}
\end{table}

\begin{table}[!t]
\begin{center}
\begin{tabular}{c|c|c|c|c|c}
\#models             & 500    & 1000    & 3000    & 7124   & 1000aug \\
\hline
MSE ($\cdot 10^{-3}$)  & $0.48$ & $0.66$ & $0.84$   & $0.97$ & $1.18$
\end{tabular}
\end{center}
\caption{Per-pixel mean squared error of image generation with varying number of car models in the training set.}
\label{tbl:quant_var_num_samples}
\end{table}

\subsection{Modeling transformations}
\begin{figure}
\begin{center}
   \includegraphics[width=1\linewidth]{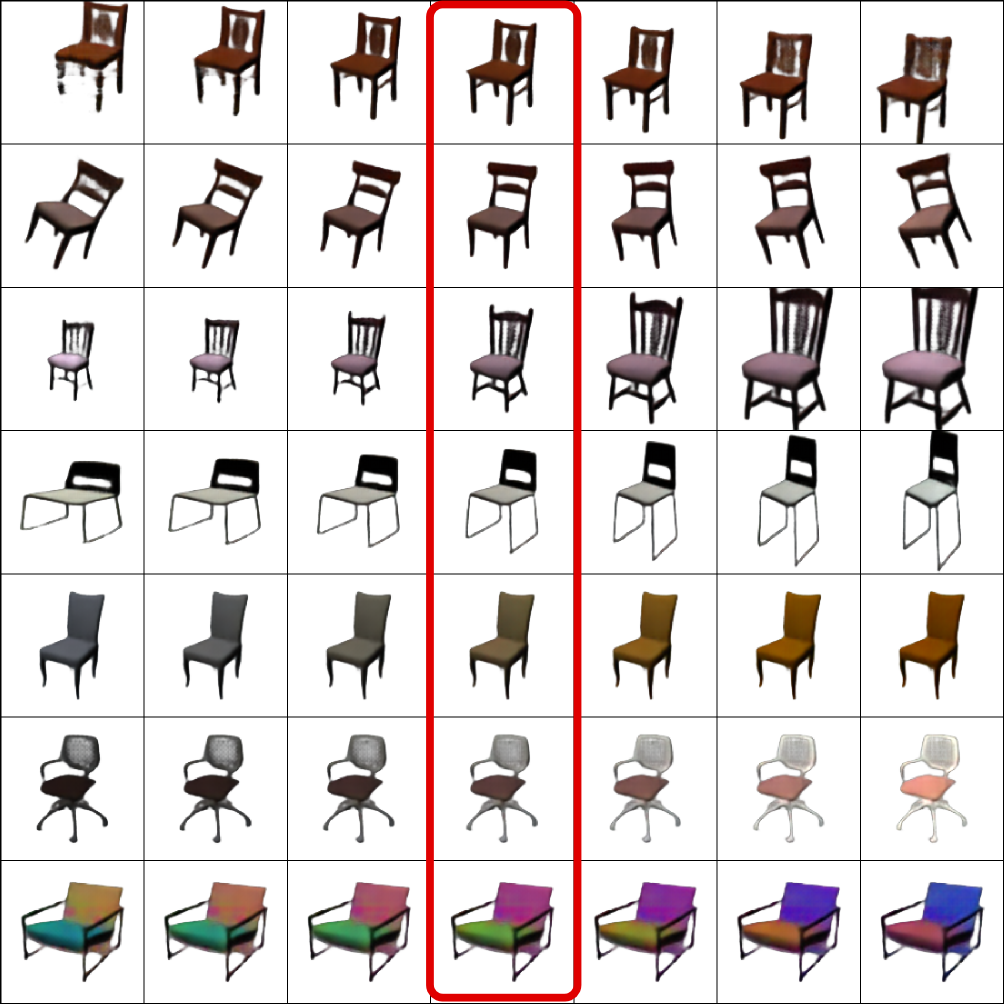}
\end{center}
   \caption{Generation of chair images while activating various transformations. Each row shows one transformation: translation, rotation, zoom, stretch, saturation, brightness, color. The middle column shows the reconstruction without any transformation.}
\label{fig:augmentation}
\end{figure}

Figure~\ref{fig:augmentation} shows how a network is able to generate chairs that are significantly transformed relative to the original images.
Each row shows a different type of transformation.
Images in the central column are non-transformed.
Even in the presence of large transformations, the quality of the generated images is basically as good as without transformation.
The image quality typically degrades a little in case of unusual chair shapes (such as rotating office chairs) and chairs including fine details such as armrests (see \eg one of the armrests in the second to last row in Figure~~\ref{fig:augmentation}).
Interestingly, the network successfully models zoom-out (row 3 of Figure~\ref{fig:augmentation}), even though it has never been presented any zoomed-out images during training.

The network easily deals with extreme color-related transformations, but has more problems representing large spatial changes, especially translations.
The generation quality in such cases could likely be improved with a more complex architecture, which would allow transformation parameters to explicitly affect the feature maps of convolutional layers (by translating, rotating, zooming them), perhaps in the fashion similar to Gregor \etal~\cite{DRAW} or Jaderberg \etal~\cite{Jaderberg_NIPS15}.

\begin{figure}
\begin{center}
\includegraphics[width=1\linewidth]{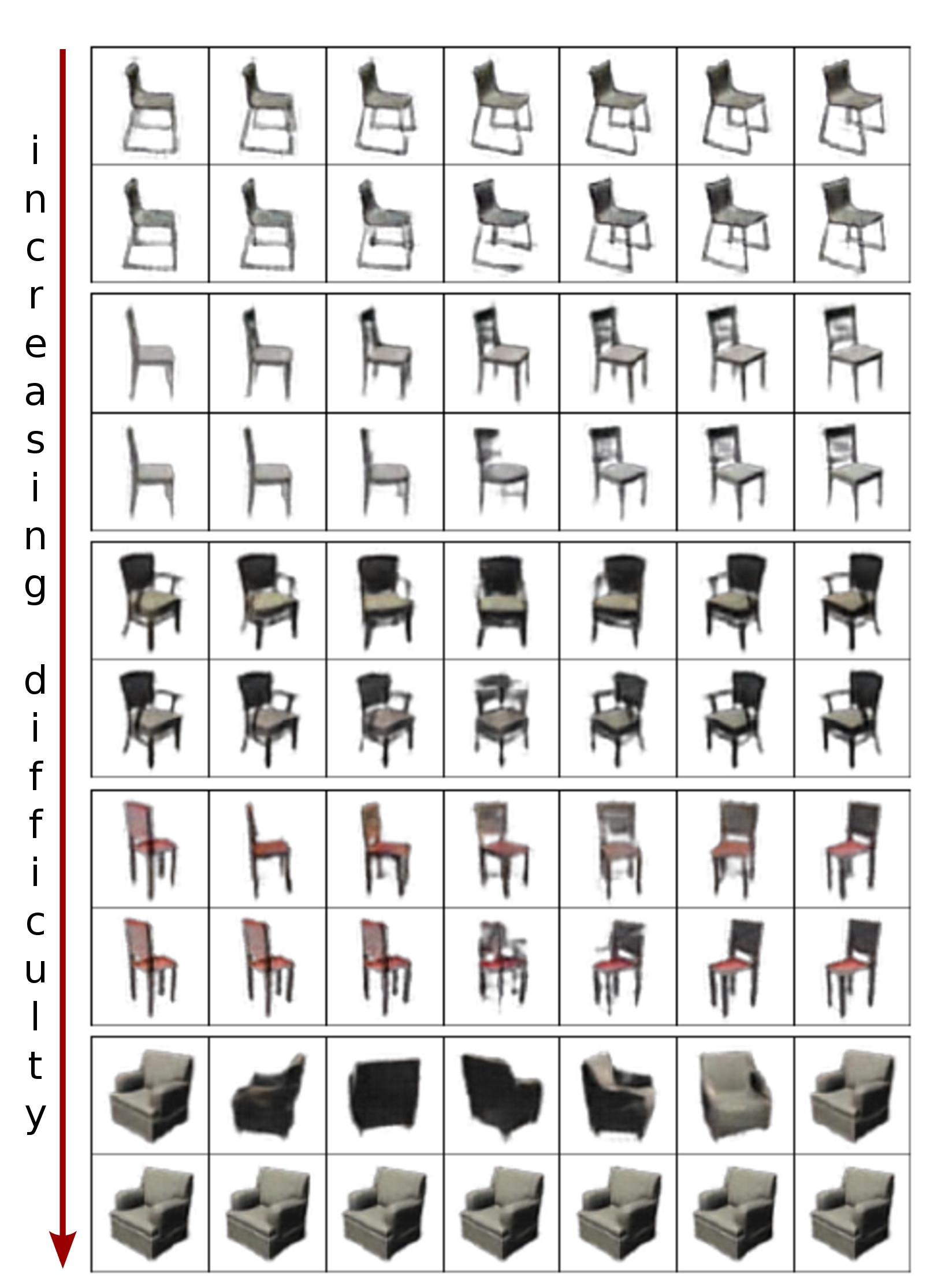}
\end{center}
   \caption{Examples of view interpolation (azimuth angle). In each pair of rows the top row is with knowledge transfer and the second row is without it. In each row the leftmost and the rightmost images are the views presented to the network during training while all intermediate ones are new to the network and, hence, are the result of interpolation. The number of different views per chair available during training is 15, 8, 4, 2, 1 (top-down). Image quality is worse than in other figures because we used the $64 \times 64$ network.}
\label{fig:angle_interpolation_examples}
\end{figure}

\begin{figure}
\begin{center}
   \includegraphics[width=1.\linewidth]{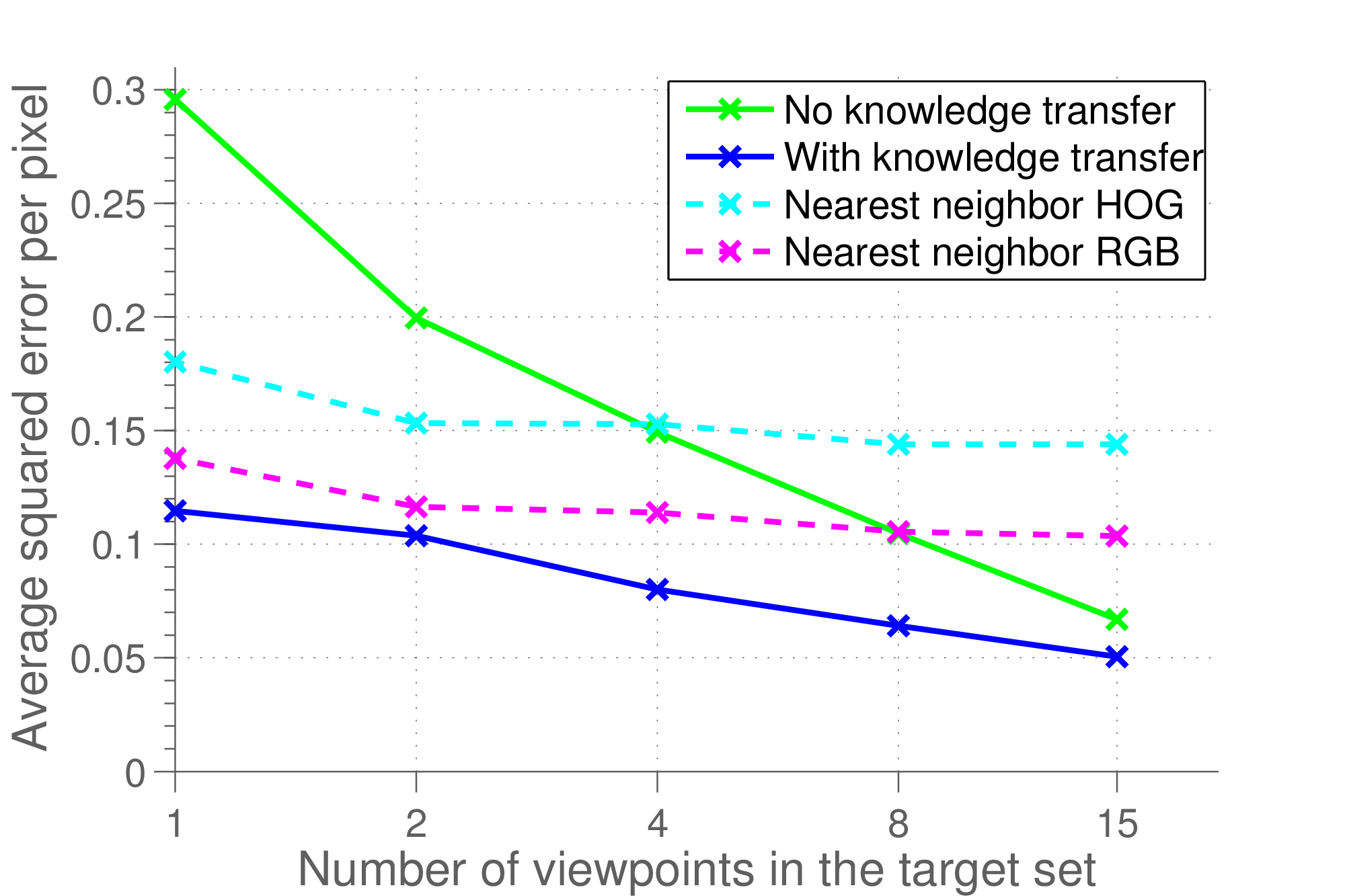}
\end{center}
   \caption{Reconstruction error for unseen views of chairs from the target set depending on the number of viewpoints present during training. Blue: all viewpoints available in the source dataset (knowledge transfer), green: the same number of viewpoints are available in the source and target datasets (no knowledge transfer).}
\label{fig:angle_interpolation_average_plot}
\end{figure}

\subsection{Interpolation between viewpoints} \label{sec:angle_interpolation}
In this section we show that the network is able to generate previously unseen views by interpolating between views present in the training data. 
This demonstrates that the network internally learns a representation of chairs which enables it to judge about chair similarity and use the known examples to generate previously unseen views.

In this experiment we used a $64 \times 64$ network to reduce computational costs. We randomly separated the chair styles into two subsets: the 'source set' with $90$ \% styles and the 'target set' with the remaining $10$ \% chairs. 
We then varied the number of viewpoints per style either in both of these subsets together ('no transfer') or just in the target set ('with transfer') and then trained a generative network as before. 
In the second setup the idea is that the network may use the knowledge about chairs learned from the source set (which includes all viewpoints) to generate the missing viewpoints of the chairs from the target set.

Figure~\ref{fig:angle_interpolation_examples} shows some representative examples of angle interpolation. For 15 views in the target set (first pair of rows) the effect of the knowledge transfer is already visible: interpolation is smoother and fine details are preserved better, for example a leg in the middle column. Starting from 8 views (second pair of rows and below) the network without knowledge transfer fails to produce satisfactory interpolation, while the one with knowledge transfer works reasonably well even with just one view presented during training (bottom pair of rows). However, in this case some fine details, such as the armrest shape, are lost.

In Figure~\ref{fig:angle_interpolation_average_plot} we plot the average squared error of the generated missing viewpoints from the target set, both with and without transfer (blue and green curves). Clearly, presence of all viewpoints in the source dataset dramatically improves the performance on the target set, especially for small numbers of available viewpoints.

One might suppose (for example looking at the bottom pair of rows of Figure~\ref{fig:angle_interpolation_examples}) that the network simply learns all the views of the chairs from the source set and then, given a limited number of views of a new chair, finds the most similar one, in some sense, among the known models and simply returns the images of that chair. To check if this is the case, we evaluated the performance of such a naive nearest neighbor approach. For each image in the target set we found the closest match in the source set for each of the given views and interpolated the missing views by linear combinations of the corresponding views of the nearest neighbors. For finding nearest neighbors we tried two similarity measures: Euclidean distance between RGB images and between HOG descriptors. The results are shown in Figure~\ref{fig:angle_interpolation_average_plot}. Interestingly, although HOG yields semantically much more meaningful nearest neighbors (not shown in figures), RGB similarity performs much better numerically. The performance of this nearest neighbor method is always worse than that of the network with knowledge transfer, suggesting that the network learns more than just linearly combining the known chairs, especially when many viewpoints are available in the target set.

\subsection{Elevation transfer and extrapolation}
\begin{figure}
\begin{center}
\begin{tabular}{c}
  \hskip .8em 0\degree \hskip 2.3em 10\degree \hskip 2.3em 20\degree \hskip 2.3em 30\degree \hskip 2.3em 40\degree \hskip 2.3em 50\degree \hskip 2.3em 70\degree \\
   \includegraphics[width=0.9\linewidth]{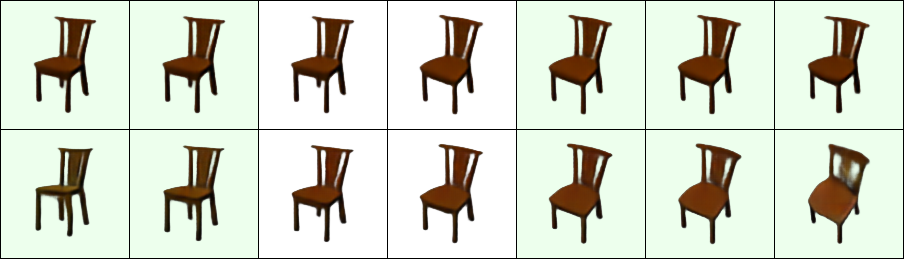}\\
   \includegraphics[width=0.9\linewidth]{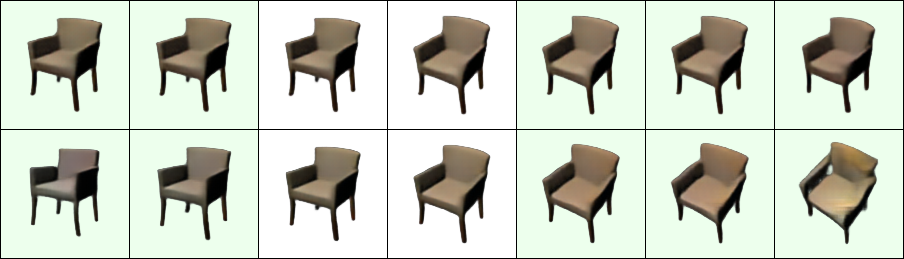}
\end{tabular}
\end{center}
   \caption{Elevation angle knowledge transfer.
   In each pair of rows \textbf{top row:} trained only on chairs (no knowledge transfer), \textbf{bottom row:} trained both on chairs and tables (with knowledge transfer).
   Green background denotes elevations not presented during training.}
\label{fig:elevation_transfer}
\end{figure}

The chairs dataset only contains renderings with elevation angles $20\degree$ and $30\degree$, while for tables elevations between $0\degree$ and $40\degree$ are available.
We show that we can transfer information about elevations from one class to another.
To this end we trained a network on both chairs and tables, and then generated images of chairs with elevations not present during training.
As a baseline we use a network trained solely on chairs.
The results are shown in Figure~\ref{fig:elevation_transfer}.
While the network trained only on chairs does not generalize to unseen elevation angles almost at all, the one trained with tables is able to generate unseen views of chairs very well.
The only drawback is that the generated images do not always precisely correspond to the desired elevation, for example $0\degree$ and $10\degree$ for the second model in Figure~\ref{fig:elevation_transfer}.
Still, this result suggests that the network is able to transfer the understanding of 3D object structure from one object class to another.

\begin{figure}
\begin{center}
\renewcommand{\arraystretch}{0.1}
\begin{tabular}{c}
  \hskip 0.6em 0\degree \hskip 1.8em 10\degree \hskip 1.8em 20\degree \hskip 1.9em 30\degree \hskip 1.8em 40\degree \hskip 1.8em 50\degree \hskip 1.8em 70\degree \hskip 1.9em 90\degree \\
  \cr\, \\
   \includegraphics[width=0.9\linewidth]{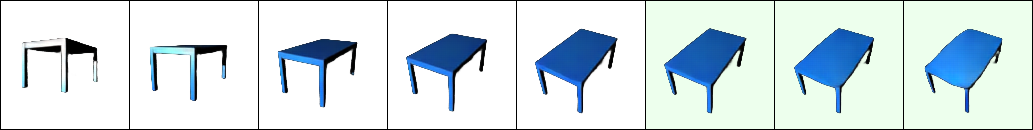}\\
   \includegraphics[width=0.9\linewidth]{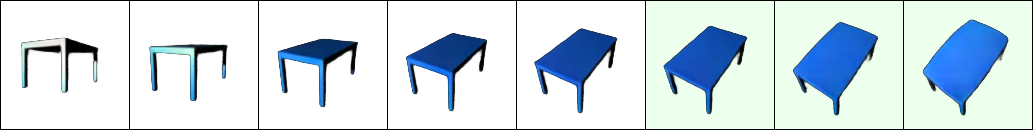}\\
   \cr\, \\
   \includegraphics[width=0.9\linewidth]{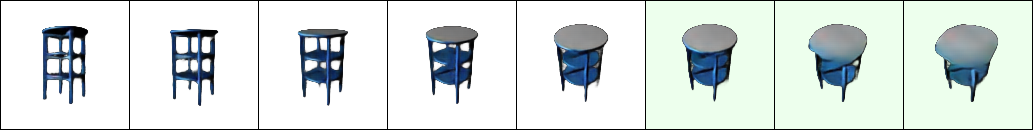}\\
   \includegraphics[width=0.9\linewidth]{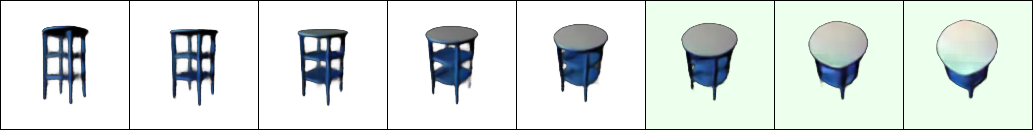}
\end{tabular}
\end{center}
   \caption{Elevation extrapolation.
   In each pair of rows \textbf{top row:} trained only on tables, \textbf{bottom row:} trained both on chairs and tables.
   Green background denotes elevations not presented during training.}
\label{fig:elevation_extrapolate}
\end{figure}

The network trained both on chairs and tables can fairly well predict views of tables from previously unseen elevation angles.
Figure~\ref{fig:elevation_extrapolate} shows how the network can generate images with previously unseen elevations from $50\degree$ to $90\degree$.
Interestingly, the presence of chairs in the training set helps better extrapolate views of tables.
Our hypothesis is that the network trained on both object classes is forced to not only model one kind of objects, but also the general 3D geometry.
This helps generating reasonable views from new elevation angles.
We hypothesize that modeling even more object classes with a single network would allow to learn a universal class-independent representation of 3D shapes.

\subsection{Interpolation between styles}
\begin{figure}
\begin{center}
   \includegraphics[width=1\linewidth]{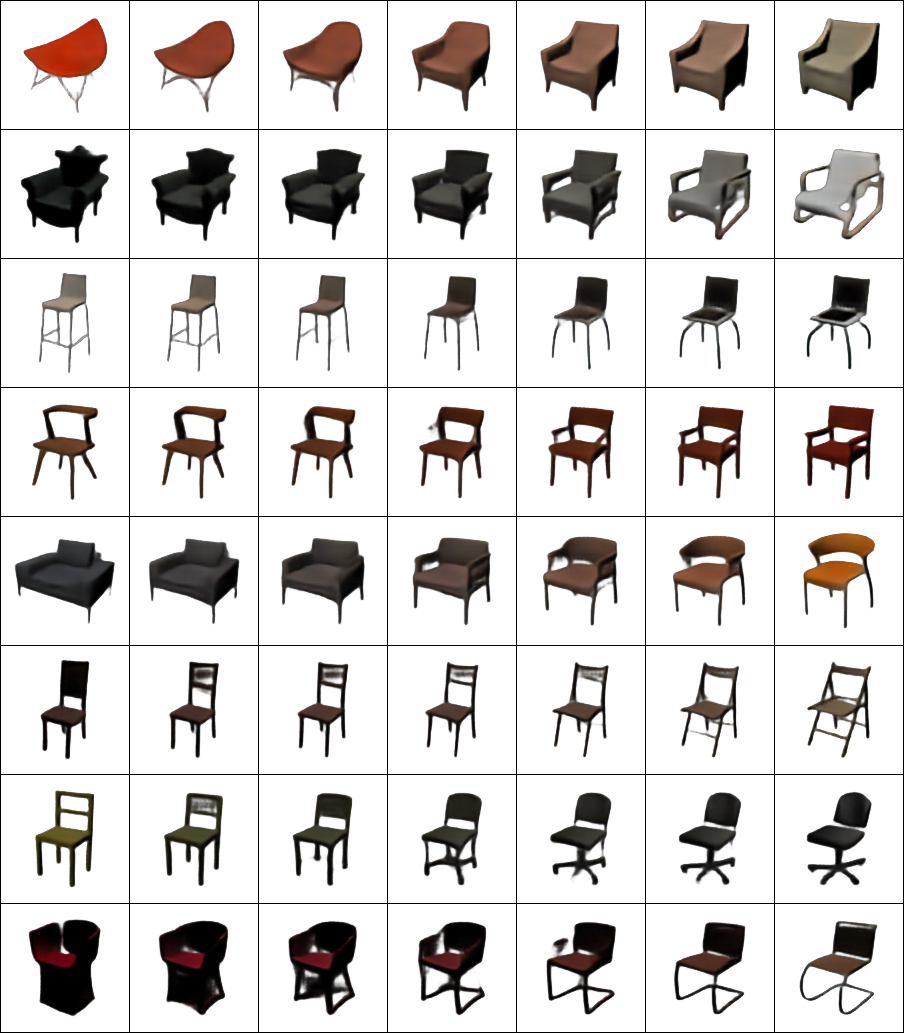}
\end{center}
   \caption{Examples of morphing different chairs, one morphing per row. Leftmost and rightmost chairs in each row are present in the training set, all intermediate ones are ``invented'' by the network. Rows are ordered by decreasing subjective quality of the morphing, from top to bottom.}
\label{fig:morphing_chairs}
\end{figure}

\begin{figure}
\begin{center}
   \includegraphics[width=1\linewidth]{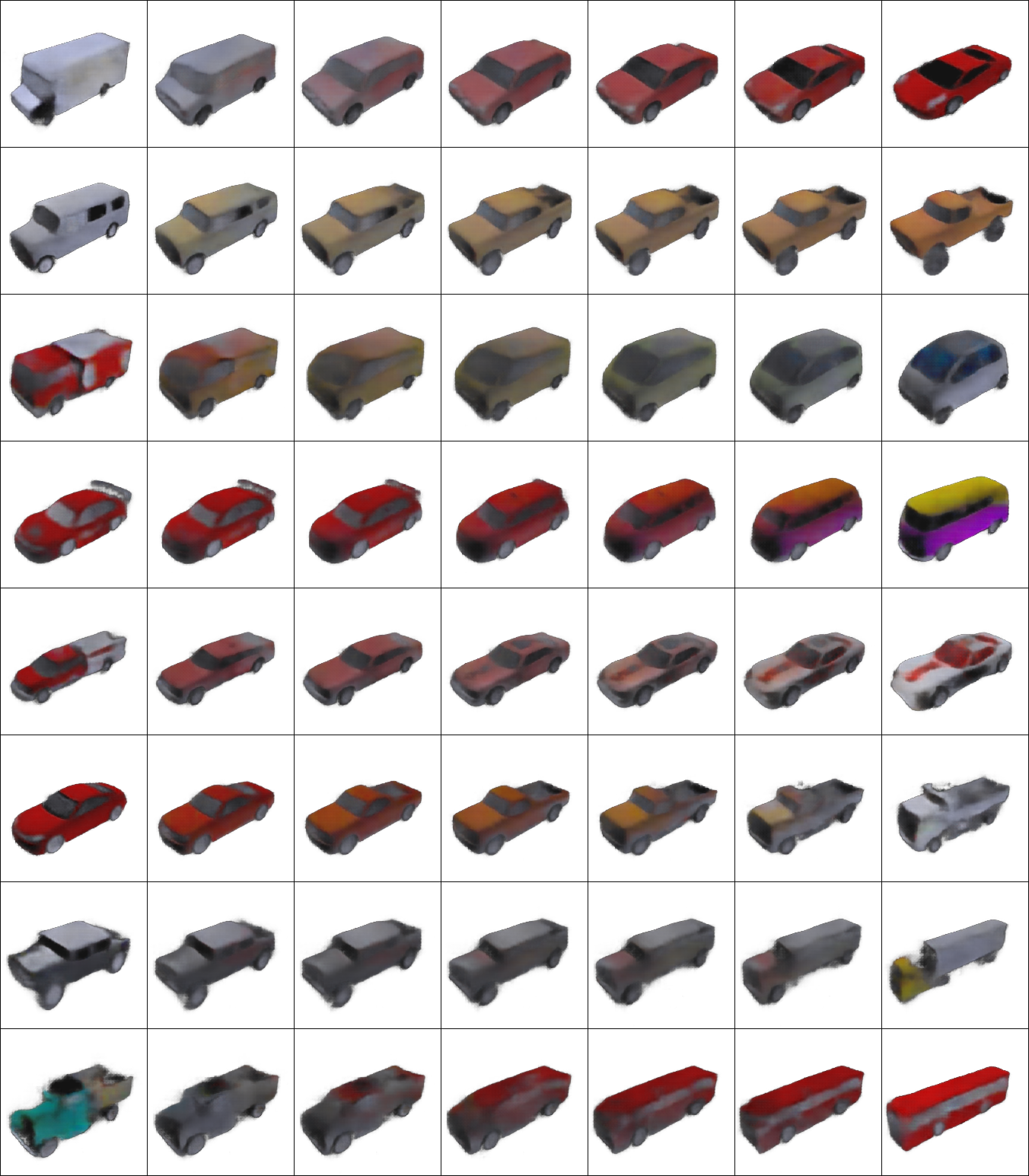}
\end{center}
   \caption{Examples of morphing different cars, one morphing per row. Leftmost and rightmost chairs in each row are present in the training set, all intermediate ones are ``invented'' by the network. Rows are ordered by decreasing subjective quality of the morphing, from top to bottom.}
\label{fig:morphing_cars}
\end{figure}

Remarkably, the generative network can not only imagine previously unseen views of a given object, but also invent new objects by interpolating between given ones.
To obtain such interpolations, we simply linearly change the input label vector from one class to another.
Some representative examples of such morphings for chairs and cars are shown in Figures~\ref{fig:morphing_chairs} and ~\ref{fig:morphing_cars} respectively.
The morphings of each object class are sorted by subjective morphing quality (decreasing from top to bottom).
The networks produce very naturally looking morphings even in challenging cases.
Even when the two morphed objects are very dissimilar, for example the last two rows in Figure~\ref{fig:morphing_chairs}, the intermediate chairs look very realistic.

\begin{figure}
\begin{center}
\includegraphics[width=1\linewidth]{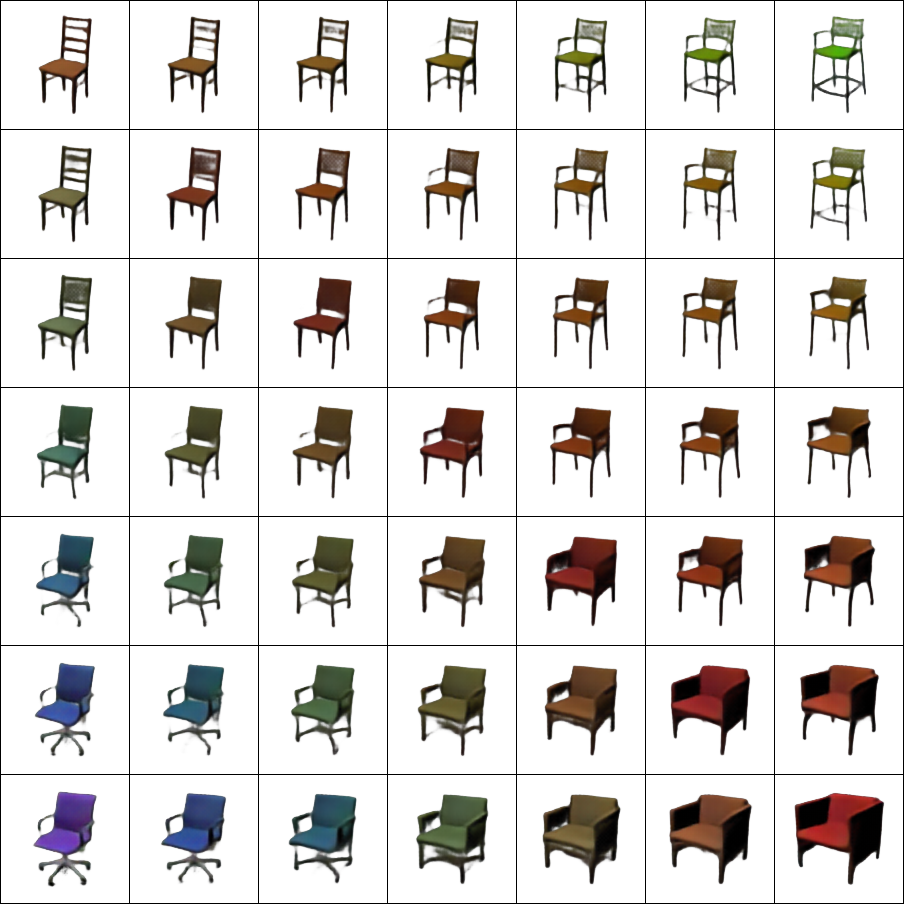}
\end{center}
   \caption{Interpolation between triples of chairs: one triple in the upper triangle of the table and one in the lower triangle. Models in the corners are present in the training set, all other images are ``invented'' by the network. }
\label{fig:morphing_three_chairs}
\end{figure}

It is also possible to interpolate between more than two objects.
Figure~\ref{fig:morphing_three_chairs} shows morphing between three chairs: one triple in the upper triangle and one in the lower triangle of the table.
The network successfully combines the features of three chairs.

\begin{figure}
\begin{center}
\includegraphics[width=1\linewidth]{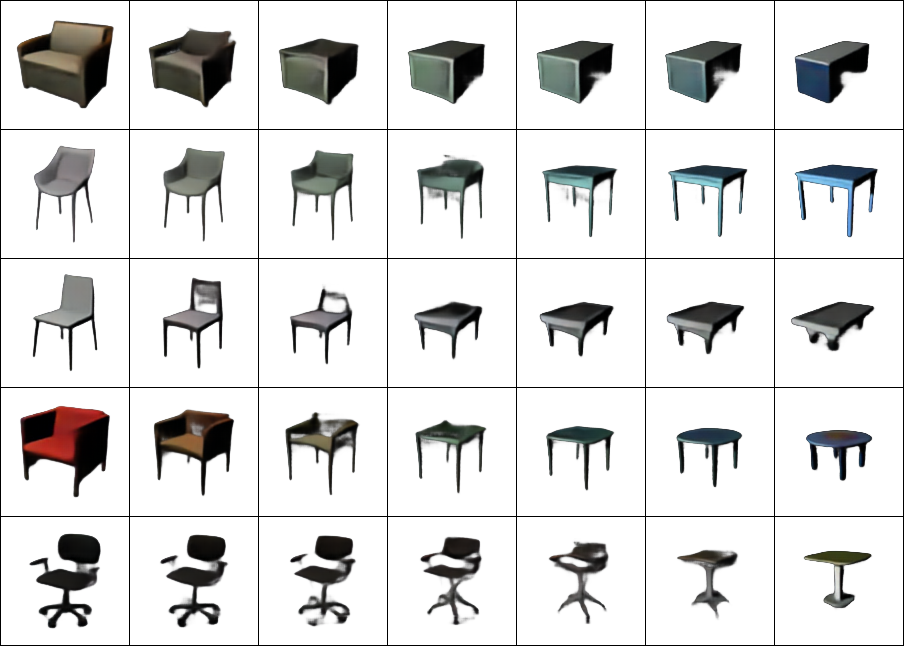}
\end{center}
   \caption{Interpolation between chairs and tables. The chairs on the left and the tables on the right are present in the training set.}
\label{fig:morphing_chairs_tables}
\end{figure}

The networks can interpolate between objects of the same class, but can they morph objects of different classes into each other?
The inter-class difference is larger than the intra-class variance, hence to successfully interpolate between classes the network has to close this large gap between different classes.
We check if a network trained on chairs and tables is capable of doing this.
Results are shown in Figure~\ref{fig:morphing_chairs_tables}.
The quality of intermediate images is slightly worse than for intra-class morphings shown above, but overall very good, especially considering that during training the network has not seen anything intermediate between a chair and a table.

\subsection{Feature space arithmetics}
\begin{figure}
\begin{center}
\includegraphics[width=1\linewidth]{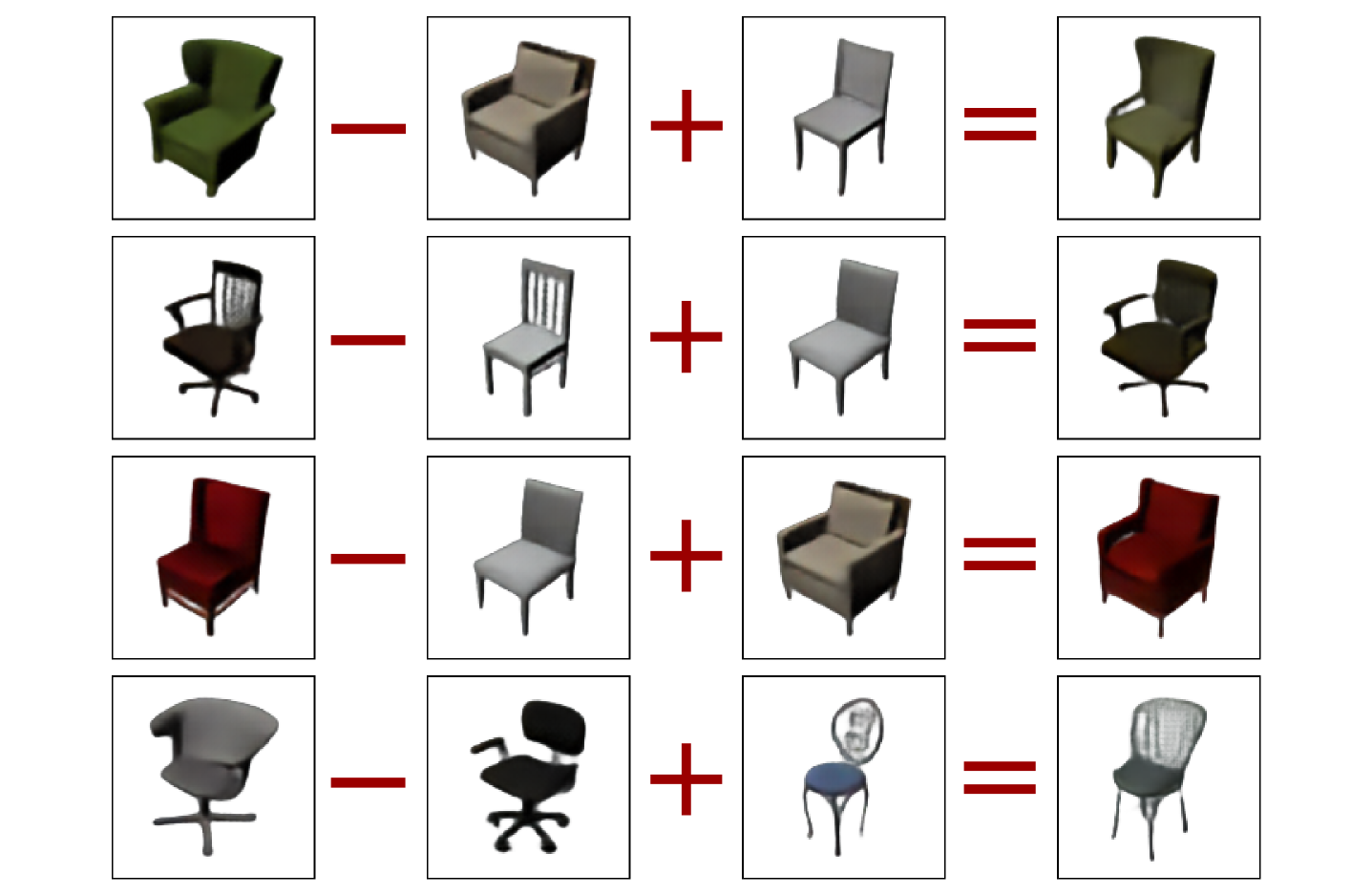}
\end{center}
   \caption{Feature arithmetics: simple operations in the feature space lead to interpretable changes in the image space.}
\label{fig:feature_arithmetic}
\end{figure}

In the previous section we have seen that the feature representation learned by the network allows for smooth transitions between two or even three different objects.
Can this property be used to transfer properties of one object onto another by performing simple arithmetics in the feature space?
Figure~\ref{fig:feature_arithmetic} shows that this is indeed possible.
By simple subtraction and addition in the feature space (FC2 features in this case) we can change an armchair into a chair with similar style, or a chair with a stick back into an identical chair with a solid back.
We found that the exact layer where the arithmetic is performed does not matter: the results are basically identical when we manipulate the input style vectors, or the outputs of layers FC1 or FC2.

\begin{figure}
\begin{center}
\setlength{\tabcolsep}{0.05cm}
\renewcommand{\arraystretch}{0.5}
\begin{tabular}{CD}
   (a) & \includegraphics[width=0.9\linewidth]{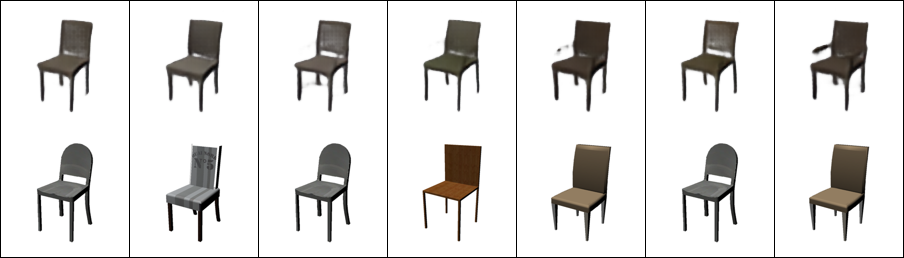}\\
   (b) & \includegraphics[width=0.9\linewidth]{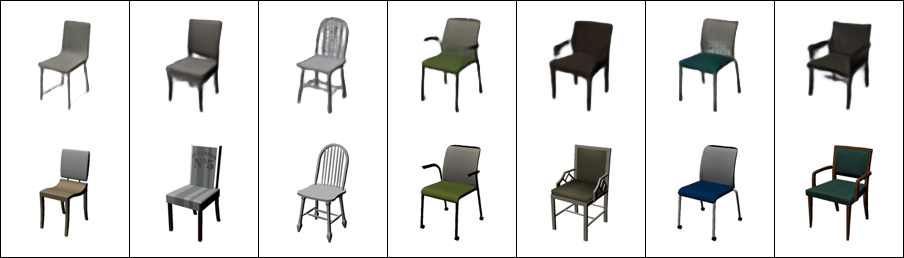}\\
   \cr\, & \\
   (c) & \includegraphics[width=0.9\linewidth]{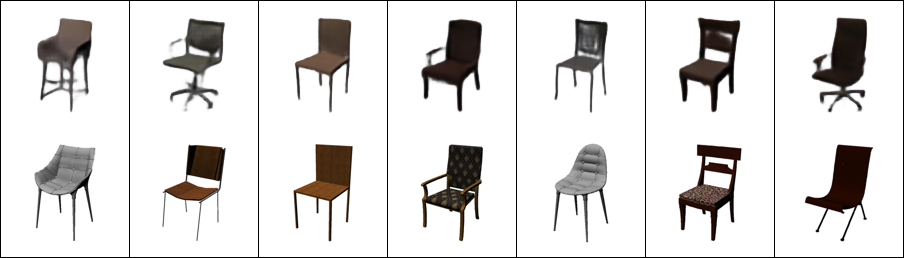}\\
   (d) & \includegraphics[width=0.9\linewidth]{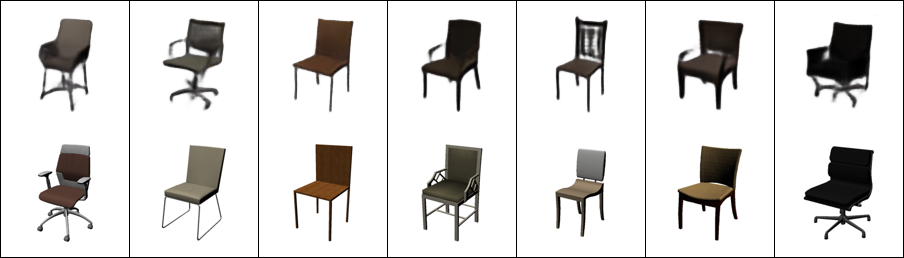}\\
   \cr\, & \\
   (e) & \includegraphics[width=0.9\linewidth]{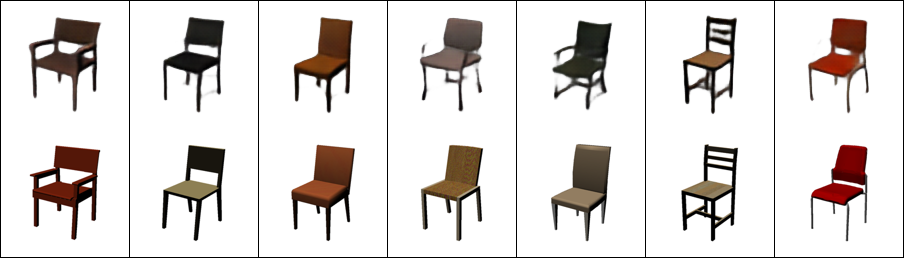}\\
   (f) & \includegraphics[width=0.9\linewidth]{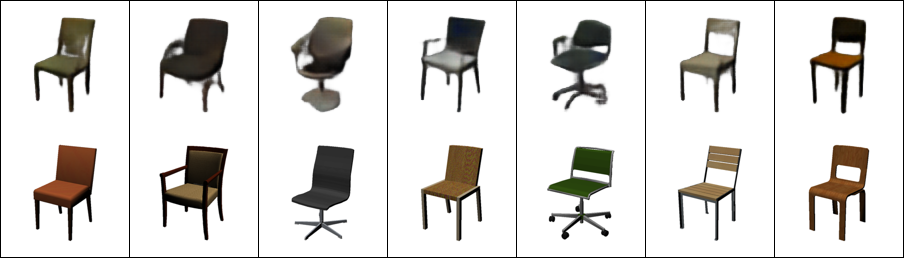}\\
   \cr\, & \\
   (g) & \includegraphics[width=0.9\linewidth]{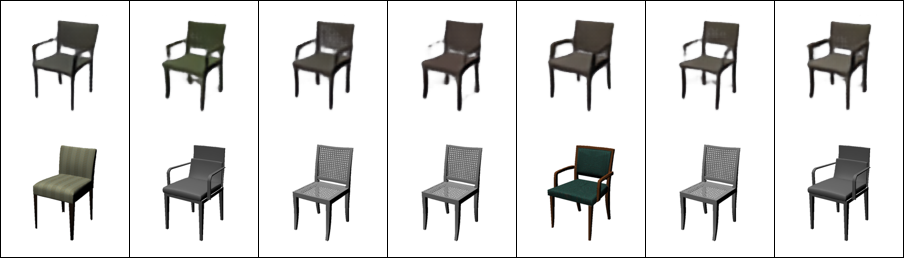}\\
   (h) & \includegraphics[width=0.9\linewidth]{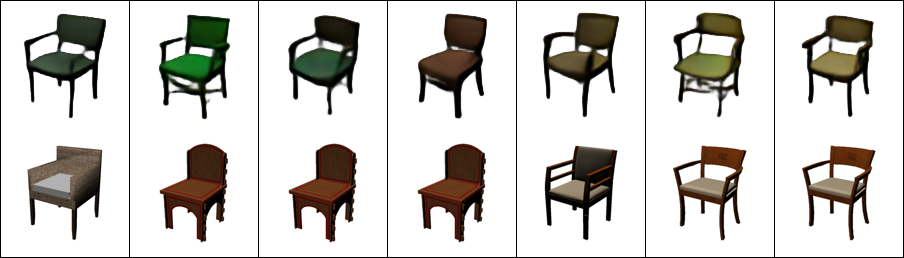}

\end{tabular}

\end{center}
   \caption{Random chairs, generated with different methods.
   In each pair of rows top: generated chairs, bottom: nearest neighbors from the training set.
   \textbf{(a),(b):} from softmax of a Gaussian in the input layer,
   \textbf{(c),(d):} interpolations between several chairs from the training set,
   \textbf{(e),(f):} from Gaussian noise in FC2 of stochastic networks trained with the variational bound loss,
   \textbf{(g),(h):} from Gaussian noise in FC2 of networks trained with usual loss.
   See the text for more details.}
\label{fig:random_chairs}
\end{figure}

\subsection{Random chair generation}
\label{sec:random_chairs}
In this section we show results on generating random chair images using the ideas briefly outlined in Section \ref{sec:prob_gen}.
In particular, we experiment both with networks trained in a fully supervised manner using the training objective from Section \ref{sec:model}, and with networks trained with the variational bound objective described in  Appendix A.

As mentioned above, there is no principled way to perform sampling using networks trained in a supervised manner. Nonetheless there are some natural heuristics that can be used to obtain ``quasi random'' chairs.
We can first observe that the style input of the network is a probability distribution over styles, which at training time is concentrated on a single style (i.e. $\class$ is a one-hot encoding of the chair style).
However, in the interpolation experiments we have seen that the network also generates plausible images given inputs with several non-zero entries.
This suggests generating random images by using random distributions as input for the network.
We tried two families of distributions: (1) we computed the softmax of a Gaussian noise vector with the same size as $\class$, with zero mean and standard deviation $\sigma$, and (2) we first randomly selected $M$ styles, then sampled coefficient for each of them from $uniform([0,1])$, then normalized to unit sum.

Exemplary results of these two experiments are shown in Figure~\ref{fig:random_chairs} (a)-(d).
For each generated image the closest chair from the dataset, according to Euclidean distance, is shown.
(a) and (b) are generated with method (1) with $\sigma=2$ and $\sigma=4$ respectively.
The results are not good: when $\sigma$ is low the generated chairs are all similar, while with higher $\sigma$ they essentially copy chairs from the training set.
(c) and (d) are produced with method (2) with $M=3$ and $M=5$ respectively.
Here the generated chairs are quite diverse and not too similar to the chairs from the training set.

The model which was trained with a variational bound objective (as described in Appendix A) directly allows us to sample from the assumed prior (Gaussian with zero mean and unit covariance) and generate images from these draws.
That is, we simply replace the FC2 activations of the style stream by random Gaussian noise.
The results are shown in Figure~\ref{fig:random_chairs} (e)-(f).
The difference between (e) and (f) is that in (e) the KL-divergence term in the loss function was weighted $10$ times higher than in (f).
This leads to much more diverse chairs being generated.

As a control, in Figure~\ref{fig:random_chairs} (g)-(h) we also show chairs generated in the same way (but with adjusted standard deviations) from a network trained without the variational bound objective.
While such a procedure is not guaranteed to result in any visually appealing images, since the hidden layer activations are not restricted to a particular regime during training, we found that it does result in sharp chair images.
However, both for low (g) and high (h) standard deviations the generated chairs are not very diverse.

Overall, the heuristics with combining several chairs and the variational-bound-based training lead to generating images of roughly similar quality and diversity.
However, the second approach is advantageous in that it allows generating images simply from a Gaussian distribution and it is more principled, potentially promising further improvement when better optimized or combined with other kinds of stochastic networks.

\subsection{Correspondences}
The ability of the generative CNN to interpolate between different chairs allows us to find dense correspondences between different object instances, even if their appearance is very dissimilar.

Given two chairs from the training dataset, we used the ``1s-S-deep'' network to generate a morphing consisting of $64$ images (with fixed view). We then computed the optical flow in the resulting image sequence using the code of Brox \etal~\cite{Brox_ECCV2004}. To compensate for the drift, we refined the computed optical flow by recomputing it with a step of 9 frames, initialized by concatenated per-frame flows. Concatenation of these refined optical flows gives the global vector field that connects corresponding points in the two chair images.

\begin{figure}
\begin{center}
\begin{tabular}{cc}
   \includegraphics[width=0.4\linewidth]{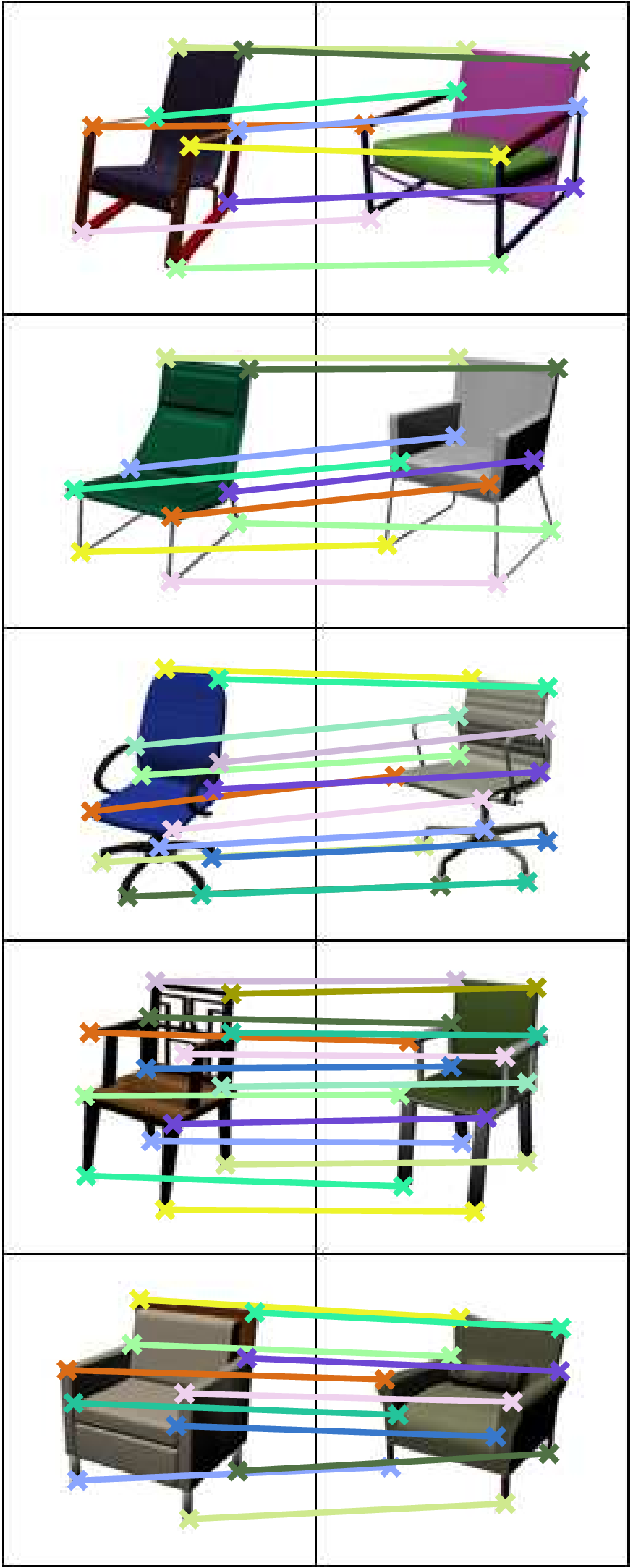} &
   \includegraphics[width=0.4\linewidth]{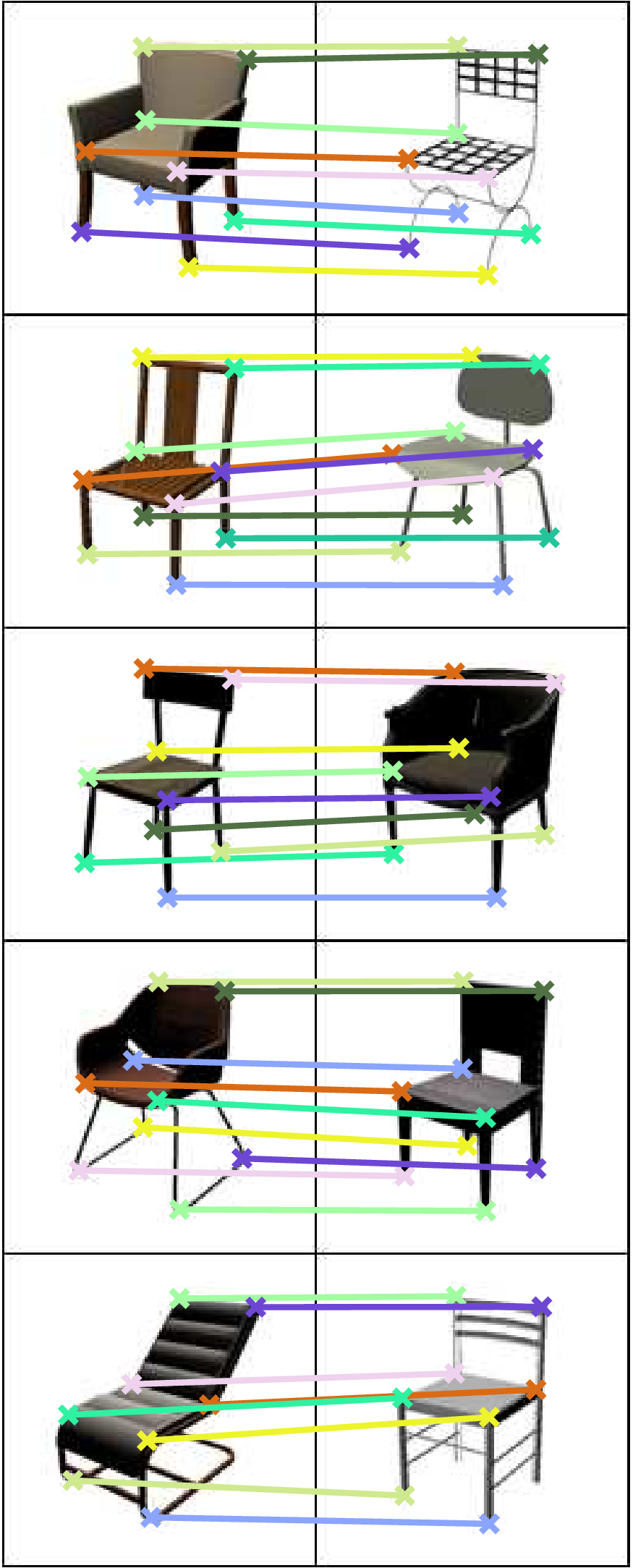} \\
\end{tabular}
\end{center}
   \caption{Exemplar image pairs from the test set with ground truth correspondences. \textbf{Left:} 'simple' pairs, \textbf{right:} 'difficult' pairs}
\label{fig:corr_test_set}
\end{figure}

In order to quantitatively evaluate the quality of the correspondences, we created a small test set of $30$ image pairs.
To analyze the performance in more detail, we separated these into 10 'simple' pairs (two chairs are quite similar in appearance) and 20 'difficult' pairs (two chairs differ a lot in appearance).
Exemplar pairs are shown in Figure~\ref{fig:corr_test_set}\,.
We manually annotated several keypoints in the first image of each pair (in total $295$ keypoints in all images) and asked $9$ people to manually mark corresponding points in the second image of each pair. We then used mean keypoint positions in the second images as ground truth. At test time we measured the performance of different methods by computing average displacement of predicted keypoints in the second images given keypoints in the first images. We also manually annotated an additional validation set of 20 image pairs to tune the parameters of all methods (however, we were not able to search the parameters exhaustively because some methods have many).

In Table~\ref{tbl:correspondence} we show the performance of our algorithm compared to human performance and two baselines: SIFT flow~\cite{Liu_ECCV2008} and Deformable Spatial Pyramid~\cite{Kim_CVPR2013} (DSP).
On average the very basic approach we used outperforms both baselines thanks to the intermediate samples produced by the generative neural network. More interestingly, while SIFT flow and DSP have problems with the difficult pairs, our algorithm does not. This suggests that errors of our method are largely due to contrast changes and drift in the optical flow, which does not depend on the difficulty of the image pair. The approaches are hence complementary: while for similar objects direct matching is fairly accurate, for more dissimilar ones intermediate morphings are very helpful.

\begin{table}
\begin{center}
\begin{tabular}{|l|c|c|c|}
\hline
Method                         & All   & Simple & Difficult\\
\hline\hline
DSP~\cite{Kim_CVPR2013}        & $5.2$          & $3.3$           & $6.3$          \\
SIFT flow~\cite{Liu_ECCV2008}  & $4.0$          & $\mathbf{2.8}$  & $4.8$          \\
Ours                           & $\mathbf{3.4}$ & $3.1$           & $\mathbf{3.5}$ \\
Human                          & $1.1$          & $1.1$           & $1.1$          \\
\hline
\end{tabular}
\end{center}
\caption{Average displacement (in pixels) of keypoints predicted by different methods on the whole test set and on the 'simple' and 'difficult' subsets.}
\vspace{-0.2cm}
\label{tbl:correspondence}
\end{table}

\section{Analysis of the network} \label{sec:analysis}
We have shown that the networks can model objects extremely well.
We now analyze the inner workings of networks, trying to get some insight into the source of their success.
The ``2s-E'' network was used in this section.

\subsection{Activating single units}
One way to analyze a neural network (artificial or real) is to visualize the effect of single neuron activations.
Although this method does not allow us to judge about the network's actual functioning, which involves a clever combination of many neurons, it still gives a rough idea of what kind of representation is created by the different network layers.

\begin{figure}
\begin{center}
\begin{tabular}{c}
   \includegraphics[width=0.97\linewidth]{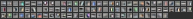} \\
   \includegraphics[width=0.97\linewidth]{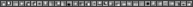}
\end{tabular}
\end{center}
   \caption{Output layer filters of the ``2s-E'' network. \textbf{Top:} RGB stream. \textbf{Bottom:} Segmentation stream.}
\label{fig:first_layer_filters}
\end{figure}

Activating single neurons of upconv3 feature maps (last feature maps before the output) is equivalent to simply looking at the filters of these layers which are shown in Figure~\ref{fig:first_layer_filters}\,. The final output of the network at each position is a linear combination of these filters. As to be expected, they include edges and blobs.

\begin{figure}
\begin{center}
\begin{tabular}{c}
   \includegraphics[width=0.9\linewidth]{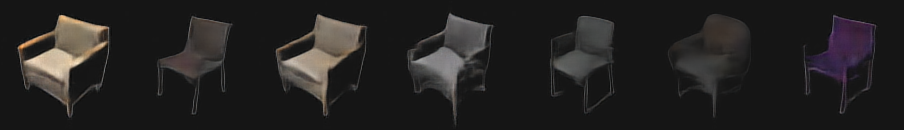}\\
   \includegraphics[width=0.9\linewidth]{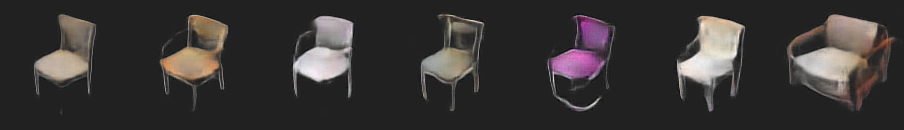}\\
   \includegraphics[width=0.9\linewidth]{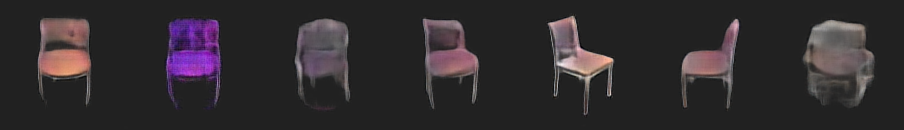}\\
   \includegraphics[width=0.9\linewidth]{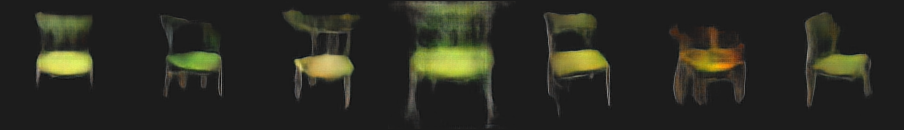}
\end{tabular}
\end{center}
   \caption{Images generated from single unit activations in feature maps of different fully connected layers of the ``2s-E'' network. \textbf{From top to bottom:} FC1 and FC2 of the class stream, FC3, FC4.}
\label{fig:single_neurons_fc}
\end{figure}

\begin{figure}[t]
\begin{center}
\begin{tabular}{c}
   \includegraphics[width=0.9\linewidth]{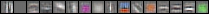}\\
   \includegraphics[width=0.9\linewidth]{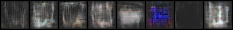}\\
   \includegraphics[width=0.9\linewidth]{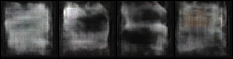}\\
\end{tabular}
\end{center}
   \caption{Images generated from single neuron activations in feature maps of some layers of the ``2s-E'' network. {From top to bottom:} upconv2, upconv1, FC5 of the RGB stream. Relative scale of the images is correct. Bottom images are $57 \times 57$ pixel, approximately half of the chair size.}
\label{fig:single_neurons_deconv}
\end{figure}

Our model is tailored to generate images from high-level neuron activations, which allows us to activate a single neuron in some of the higher layers and forward-propagate down to the image.
The results of this procedure for different layers of the network are shown in Figures~\ref{fig:single_neurons_fc} and~\ref{fig:single_neurons_deconv}.
Each row corresponds to a different network layer.
The leftmost image in each row is generated by setting all neurons of the layer to zero, and the other images~ -- by activating one randomly selected neuron.

\begin{figure}
\begin{center}
   \includegraphics[width=0.9\linewidth]{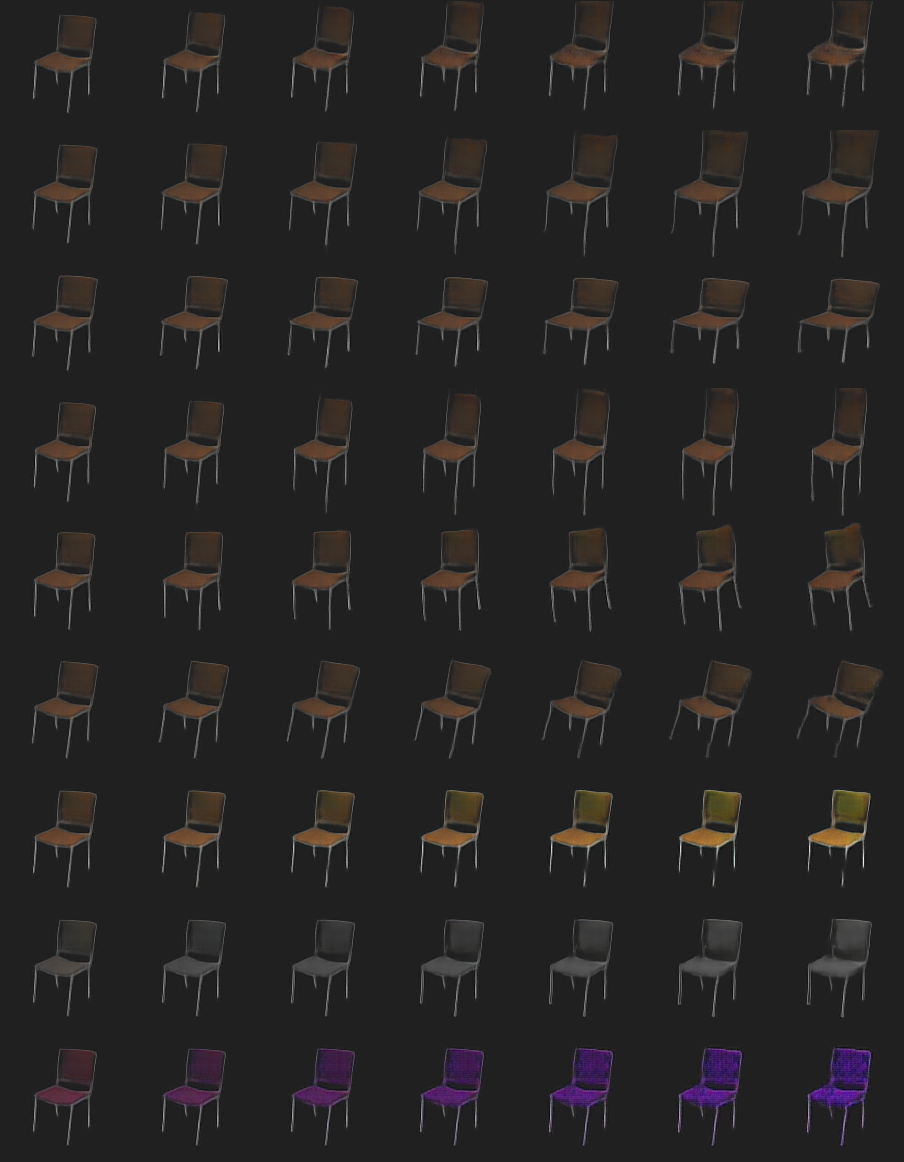}
\end{center}
   \caption{The effect of specialized neurons in the layer FC4. Each row shows the result of increasing the value of a single FC4 neuron given the feature maps of a real chair. Effects of all neurons, top to bottom: translation upwards, zoom, stretch horizontally, stretch vertically, rotate counter-clockwise, rotate clockwise, increase saturation, decrease saturation, make violet.}
\label{fig:specialized_neurons}
\end{figure}

\begin{figure}
\begin{center}
\begin{tabular}{c}
   \includegraphics[width=1\linewidth]{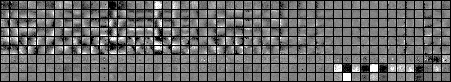}\\
\end{tabular}
\end{center}
   \caption{Some of the neural network weights corresponding to the transformation neurons shown in Figure~\ref{fig:specialized_neurons}. Each row shows weights connected to one neuron, in the same order as in Figure~\ref{fig:specialized_neurons}. Only a selected subset of most interesting channels is shown.}
\label{fig:specialized_neurons_weights}
\end{figure}

In Figure~\ref{fig:single_neurons_fc} the first two rows show images produced when activating neurons of FC1 and FC2 feature maps of the class stream while keeping viewpoint and transformation inputs fixed.
The results clearly look chair-like but do not show much variation (the most visible difference is chair vs armchair), which suggests that larger variations are achievable by activating multiple neurons.
The last two rows show results of activating neurons of FC3 and FC4 feature maps.
These feature maps contain joint class-viewpoint-transformation representations, hence the viewpoint is not fixed anymore.
The generated images still resemble chairs but get much less realistic.
This is to be expected: the further away from the inputs, the less semantic meaning there is in the activations.

In the middle of the bottom row in Figure~\ref{fig:single_neurons_fc} one can notice a neuron which seems to generate a zoomed chair.
By looking at FC4 neurons more carefully we found that this is indeed a 'zoom neuron', and, moreover, for each transformation there is a specialized neuron in FC4.
The effect of these is shown in Figure~\ref{fig:specialized_neurons}.
Increasing the activation of one of these neurons while keeping all other activations in FC4 fixed results in a transformation of the generated image.
It is quite surprising that information about transformations is propagated without change through all fully connected layers.
It seems that all potential transformed versions of a chair are already contained in the FC4 features, and the 'transformation neurons' only modify them to give more relevance to the activations corresponding to the required transformation.
The corresponding weights connecting these specialized neurons to the next layer are shown in Figure~\ref{fig:specialized_neurons_weights} (one neuron per row).
Some output channels are responsible for spatial transformations of the image, while others deal with the color and brightness.

Images generated from single neurons of the convolutional layers are shown in Figure~\ref{fig:single_neurons_deconv}\,.
A somewhat disappointing observation is that while single neurons in later layers (upconv2 and upconv3) produce edge-like images, the neurons of higher deconvolutional layers generate only blurry 'clouds', as opposed to the results of Zeiler and Fergus~\cite{Zeiler_ECCV2014} with a classification network and max-unpooling.
Our explanation is that because we use naive regular-grid upsampling, the network cannot slightly shift small parts to precisely arrange them into larger meaningful structures.
Hence it must find another way to generate fine details.
In the next subsection we show that this is achieved by a combination of spatially neighboring neurons.

\subsection{Analysis of the hidden layers}
Rather than just activating single neurons while keeping all others fixed to zero, we can use the network to normally generate an image and then analyze the hidden layer activations by either looking at them or modifying them and observing the results.
An example of this approach was already used above in~Figure~\ref{fig:specialized_neurons} to understand the effect of the 'transformation neurons'.
We present two more results in this direction here.

\begin{figure}
\begin{center}
\begin{tabular}{c}
   \includegraphics[width=0.9\linewidth]{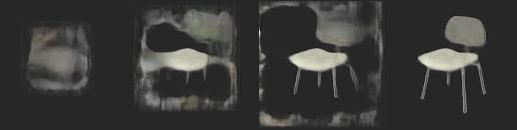}\\
\end{tabular}
\end{center}
   \caption{Chairs generated from spatially masked FC5 feature maps
     (the feature map size is $8\times 8$). The size of the non-zero
     region increases left to right: $2\times 2$, $4\times 4$,
     $6\times 6$, $8\times 8$.}
\label{fig:fc5_vary_area}
\end{figure}

First, we study how the blurry 'clouds' generated by single high-level deconvolutional neurons (Figure~\ref{fig:single_neurons_deconv}\,) form perfectly sharp chair images.
We start with the FC5 feature maps of a chair, which have a spatial extent of $8 \times 8$.
Next we only keep active neurons in a region around the center of the feature map (setting all other activations to zero), gradually increasing the size of this region from $2 \times 2$ to $8 \times 8$.
This means we go from nearly single-neuron activation level to the whole image level.
The outcome is shown in Figure~\ref{fig:fc5_vary_area}\,.
Clearly, the interaction of neighboring neurons is very important: in the central region, where many neurons are active, the image is sharp, while in the periphery it is blurry. One interesting effect that is visible in the images is how sharply the legs of the chair end in the second to last image but
appear in the larger image.
This suggests highly non-linear suppression effects between activations of neighboring neurons.

\begin{figure}
\begin{center}
\begin{tabular}{c}
   \includegraphics[width=0.8\linewidth]{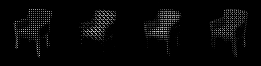}\\
   \includegraphics[width=0.8\linewidth]{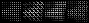}\\
   \includegraphics[width=0.8\linewidth]{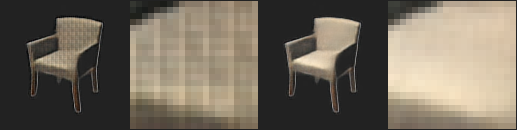}\\
\end{tabular}
\end{center}
   \caption{\textbf{Top:} Selected feature maps from the pre-output layer (upconv3) of the RGB stream. These feature maps correspond to the filters which look near-empty in Figure~\ref{fig:first_layer_filters}. \textbf{Middle:} Close-ups of the feature maps. \textbf{Bottom:} Generation of a chair with these feature maps set to zero (left image pair) or left unchanged (right). Note the high-frequency artifacts in the left pair of images.}
\label{fig:patterns}
\end{figure}

Second, interesting observations can be made by taking a closer look at the feature maps of the upconv3 layer (the last pre-output layer). Some of them exhibit regular patterns shown in Figure~\ref{fig:patterns}\,. These feature maps correspond to filters which look near-empty in Figure~\ref{fig:first_layer_filters} (such as the 3rd and 10th filters in the first row). Our explanation of these patterns is that they compensate high-frequency artifacts originating from fixed filter sizes and regular-grid upsampling. This is supported by the last row of~Figure~\ref{fig:patterns} which shows what happens to the generated image when these feature maps are set to zero.


%

\section{Conclusions}

We have shown that supervised training of convolutional neural networks can be used not only for discriminative tasks, but also for generating images given high-level style, viewpoint, and lighting information.
A network trained for such a generative task does not merely learn to generate the training samples, but learns a generic implicit representation, which allows it to smoothly morph between different object views or object instances with all intermediate images being meaningful.
Moreover, when trained in a stochastic regime, it can be creative and invent new chair styles based on random noise. 
Our experiments suggest that networks trained to reconstruct objects of two different classes develop some understanding of 3D shape and geometry.
We expect that training a single network on yet more object classes would also succeed, and constitutes an interesting direction for future work.
It is fascinating that the relatively simple architecture we proposed is already able to learn all these complex behaviors.

\ifCLASSOPTIONcompsoc
  \section*{Acknowledgments}
\else
  \section*{Acknowledgment}
\fi

AD, MT and TB acknowledge funding by the ERC Starting Grant VideoLearn (279401). JTS is supported by the BrainLinks-BrainTools Cluster of Excellence funded by the German Research Foundation (EXC 1086).

\ifCLASSOPTIONcaptionsoff
  \newpage
\fi



%

\section*{Appendix A: Training using a variational bound}
\label{sect:variational_bound}
Here we give additional details on the training objective used the experiment on random chair generation in section~\ref{sec:random_chairs}.
We phrase the problem of generating chairs as that of learning a (conditional) probabilistic generative model.
Let $\deconv_{RGB}$ and $\deconv_{segm}$ denote the ``expanding'' part of the generator and let us assume full knowledge about the view $\view$, transformation parameters $\augparams$ and chair identity $\class$. We further assume there exists a distribution $p(\bz \mid \class)$ over latent states $\bz \in \mathbb{R}^{512}$ -- which capture the underlying manifold of chair images -- such that the generators $\deconv_{RGB}$ and $\deconv_{segm}$ can generate the corresponding image using $\hat{\fc}(\bz, \view, \augparams)$ as input. We define $\hat{\fc}$ as the mapping obtained by replacing the independent processing stream for the class identity $\class$ with the 512 dimensional random vector $\bz$ in layer FC2 (c.f. Figure \ref{fig:network_architecture}).
We can then define the likelihood of a segmentation mask $\segm^i$ under transformation $T_{\augparams^i}$ as
\begin{equation}
  p(T_{\augparams^i} (\segm^i) \mid \bz^i, \augparams^i, \view^i) = \prod \deconv_{segm}\left(\hat{\fc}(\bz^i, \view^i, \augparams^i)\right),
\end{equation}
where $\deconv_{segm}$ outputs per-pixel probabilities (i.e. it has a sigmoid output nonlinearity).
Assuming the pixels in an image  $\RGB^i$ are distributed according to a multivariate Gaussian distribution the likelihood of $\RGB^i$ can be defined as
\begin{equation}
 p \left( T_{\augparams^i}(\RGB^i \cdot \segm^i) \mid \bz^i, \augparams^i, \view^i \right) = \mathcal{N} \left(\deconv_{RGB}\left(\hat{\fc}(\bz^i, \view^i, \augparams^i)\right), \bSigma \right),
\end{equation}
where $\mathcal{N}(\mu, \bSigma)$ denotes a Gaussian distribution with mean $\mu$ and covariance $\bSigma$. We simplify this formulation by assuming a diagonal covariance structure $\bSigma = \bI \sigma$.
Since all distributions appearing in our derivation are conditioned on the augmentation parameters $\augparams^i$ and view parameters $\view^i$ we will omit them in the following to simplify notation.
The marginal log likelihood of an image and its segmentation mask is then
\begin{equation}
\begin{aligned}
  \log p(\segm^i, \RGB^i) = \mathbb{E}_{\bz} \Big[ &p(T_{\augparams^i} \segm^i \mid \bz^i)  p \left( T_{\augparams^i}(\RGB^i \cdot \segm^i) \mid  \bz^i \right) \Big], \\
&\text{with } \bz^{i} \sim p(\bz \mid \class^i).
\end{aligned}
\label{eq:marginal_ll}
\end{equation}
Since we, a priori, have no knowledge regarding the structure of $p(\bz \mid \class^i)$ -- i.e. we do not know which \emph{shared} underlying structure different chairs posses -- we have to replace it with an approximate inference distribution $q(\bz \mid \class^i) = \mathcal{N} (\mu_{\bz^i}, \bI \sigma_{\bz^i})$. 
We parameterize $q$ as a two layer fully connected neural network (with 512 units each) predicting mean $\mu_{\bz^i}$ and variance $\sigma_{\bz^i}$ of the distribution. 
We refer to the parameters of this network with $\vaeparams$. Following recent examples from the neural networks literature \citep{Kingma_ICLR2014,Rezende_ICML2014} we can then jointly train this approximate inference network and the generator networks by maximizing a variational lower bound on the log-likelihood from \eqref{eq:marginal_ll} as
\begin{equation}
\begin{aligned}
  \log \ &p(\segm^i, \RGB^i) \geq \\ &\mathbb{E}_{\bz}  \Big[ \log \frac{p(T_{\augparams^i} \segm^i \mid \bz^i) p \left( T_{\augparams^i}(\RGB^i \cdot \segm^i) \mid \bz^i \right)}{q_\vaeparams(\bz \mid \class^i)} \Big] \\
 = \ & \mathbb{E}_{\bz} \Big[ \log p(T_{\augparams^i} \segm^i \mid \bz^i) + \log p \left( T_{\augparams^i}(\RGB^i \cdot \segm^i) \mid  \bz^i \right) \Big]  \\
 & - KL\Big(q(\bz \mid \class^i) \| p(\bz^i)\Big), \\
&\text{with } \bz^i \sim q_\vaeparams(\bz \mid \class^i) = \mathcal{L}^{VB}(\RGB^i, \segm^i. \class^i, \view^i, \augparams^i),
\end{aligned}
\label{eq:elbo}
\end{equation}
where $p(\bz)$ is a prior on the latent distribution, which we always set as $p(\bz) = \mathcal{N}(\mathbf{0}, \mathbf{1})$. The optimization problem we seek to solve can then be expressed through the sum of these losses over all data points (and $M$ independent samples from $q_\phi$) and is given as
\begin{equation}
\begin{aligned}
\max_{\mathbf{W}, \vaeparams} \;&\sum_{i=1}^N \sum^M_{1} \mathcal{L}^{VB}(\RGB^i, \segm^i. \class^i, \view^i, \augparams^i), \\
& \text{with } \bz \sim q_\vaeparams(\bz \mid \class^i, \view^i, \augparams^i),
\end{aligned}
\label{eq:elbo_dataset}
\end{equation}
where, for our experiments, we simply take only one sample $M = 1$ per data point.
Equation \eqref{eq:elbo_dataset} can then be optimized using standard stochastic gradient descent since the derivative with respect to all parameters can be computed in closed form. For a detailed explanation on how one can back-propagate through the sampling procedure we refer to Kingma \etal \cite{Kingma_ICLR2014}.


{\small
\bibliographystyle{IEEEtran}
\bibliography{dosovits_new}
}

%

\begin{IEEEbiography}[{\includegraphics[width=1in,height=1.25in,clip,keepaspectratio]{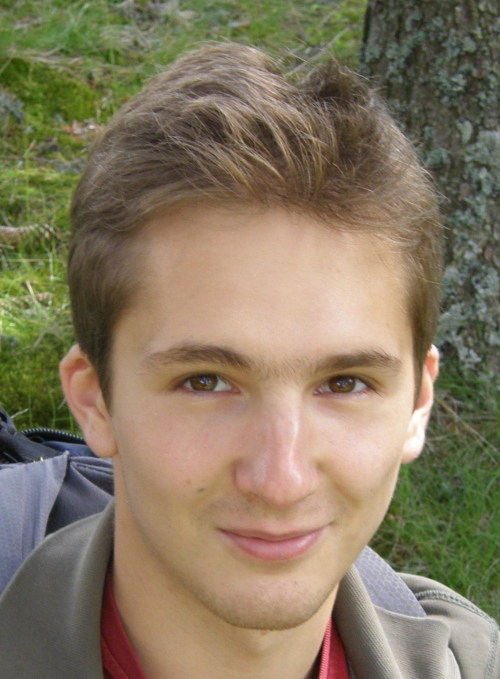}}]{Alexey Dosovitskiy} received his Specialist (equivalent of MSc, with distinction) and Ph.D. degrees in mathematics from Moscow State University in 2009 and 2012 respectively. His Ph.D. thesis is in the field of functional analysis, related to measures in infinite-dimensional spaces and representations theory. In summer 2012 he spent three months at the Computational Vision and Neuroscience Group at the University of T{\"u}bingen. Since September 2012 he is a postdoctoral researcher with the Computer Vision Group at the University of Freiburg in Germany. His current main research interests are computer vision, machine learning and optimization.
\end{IEEEbiography}

\begin{IEEEbiography}[{\includegraphics[width=1in,height=1.25in,clip,keepaspectratio]{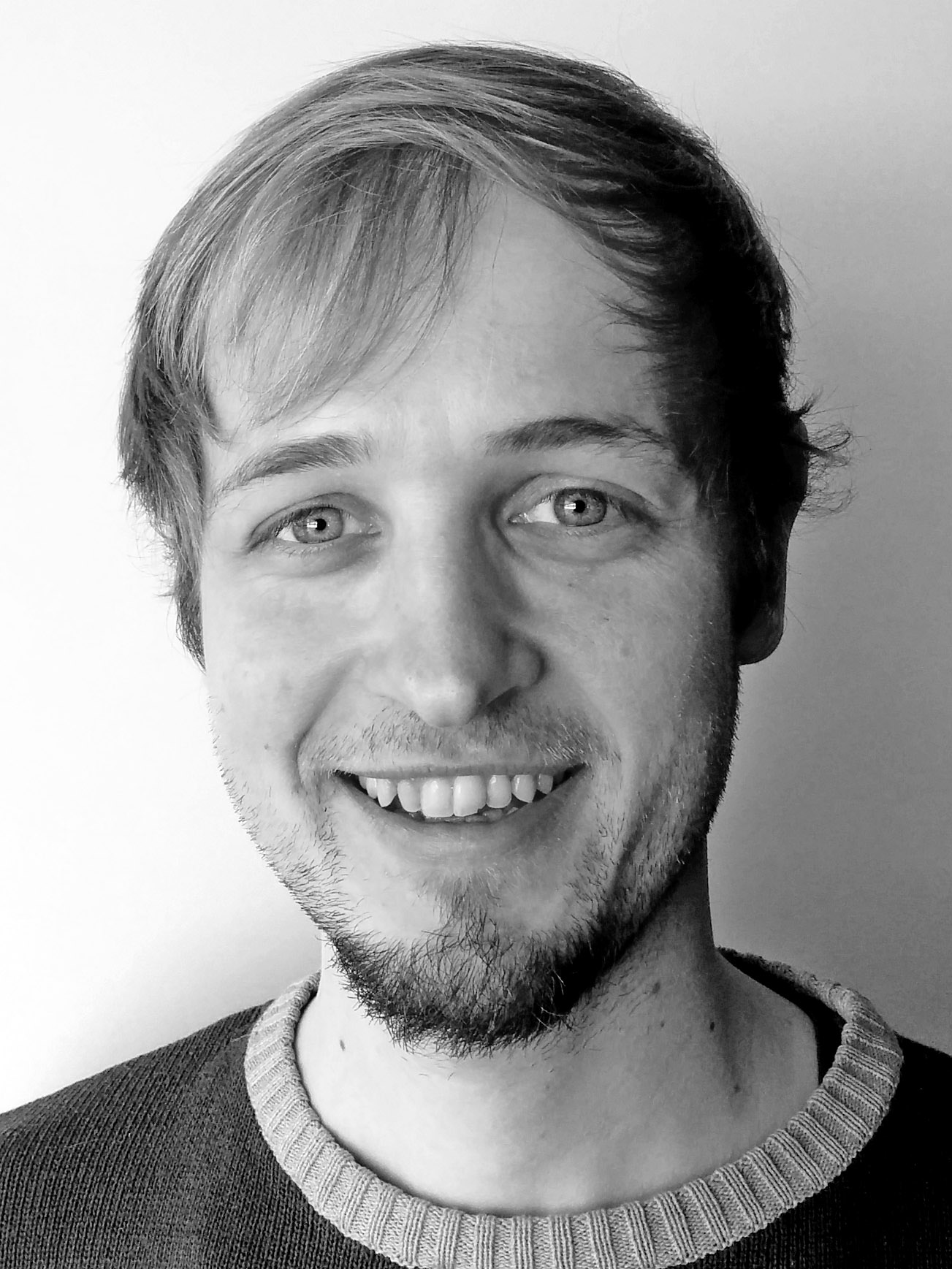}}]{Jost Tobias Springenberg}
 is a PhD student in the machine learning lab at the University of Freiburg, Germany, supervised by Martin Riedmiller. Prior to starting his PhD Tobias studied Cognitive Science at the University of Osnabrueck, earning his BSc in 2009. From 2009-2012 he then went to obtain a MSc in Computer Science from the University of Freiburg, focusing on representation learning with deep neural networks for computer vision problems. His research interests include machine learning, especially representation learning, and learning efficient control strategies for robotics.
\end{IEEEbiography}

\begin{IEEEbiography}[{\includegraphics[width=1in,height=1.25in,clip,keepaspectratio]{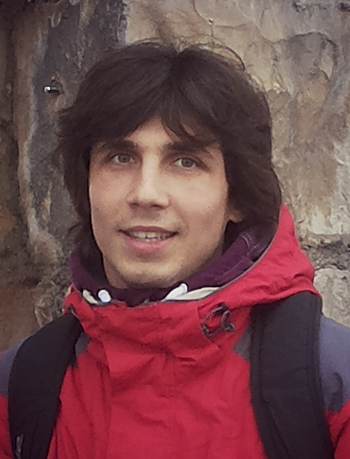}}]{Maxim Tatarchenko}
Maxim Tatarchenko received his BSc with honors in Applied Mathematics from Russian State Technological University "MATI" in 2011. He spent three years developing numerical algorithms for satellite navigation at an R\&D Company 'GPSCOM', Moscow. He proceeded with master studies at the University of Freiburg and is now close to obtaining his MSc degree. In January 2016 he is going to join the Computer Vision Group of Prof. Dr. Thomas Brox as a PhD student. His main research interest is computer vision with a special focus on deep generative models for images.

\end{IEEEbiography}

\begin{IEEEbiography}[{\includegraphics[width=1in,height=1.25in,clip,keepaspectratio]{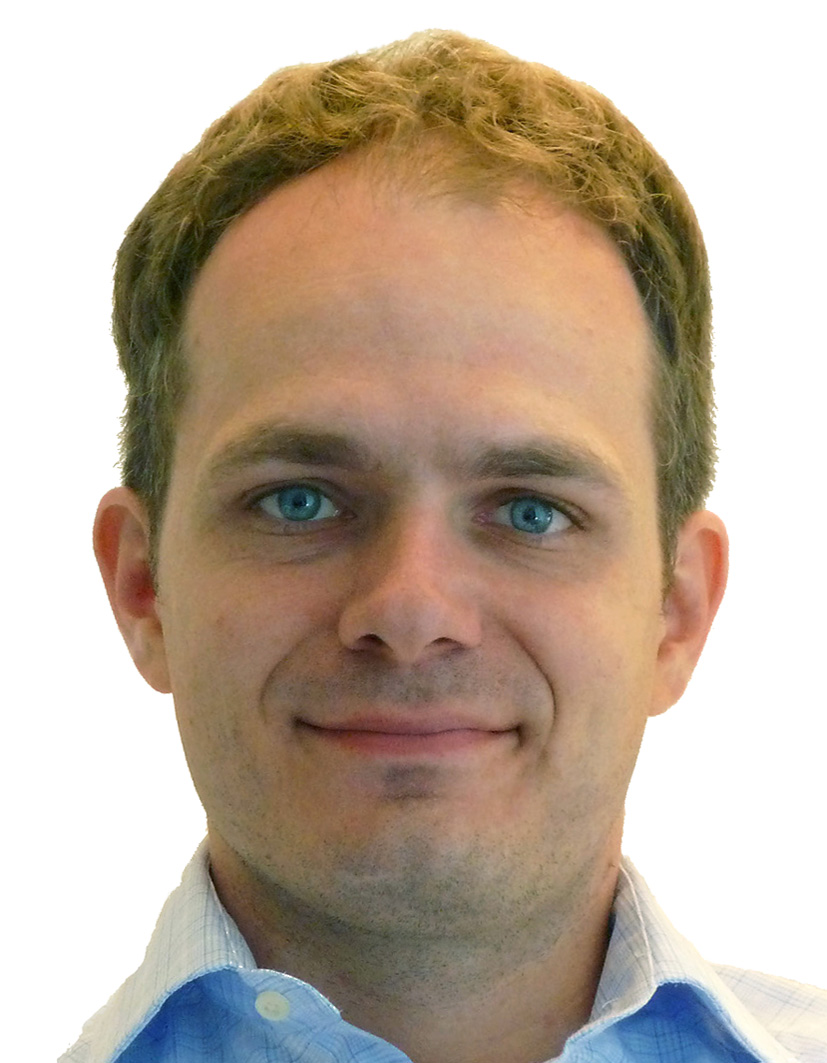}}]{Thomas Brox} received his Ph.D. degree in computer science from the Saarland University in Germany in 2005. He spent two years as a postdoctoral researcher at the University of Bonn and two years at the University of California at Berkeley. Since 2010, he is heading the Computer Vision Group at the University of Freiburg in Germany. His research interests are in computer vision, in particular video analysis and deep learning for computer vision. Prof. Brox is associate editor of the IEEE Transactions on Pattern Analysis and Machine Intelligence and the International Journal of Computer Vision. He received the Longuet-Higgins Best Paper Award and the Koendrink Prize for Fundamental Contributions in Computer Vision.
\end{IEEEbiography}






\end{document}